\documentclass[Afour,sageh,times]{sagej}

\usepackage{moreverb,url}

\usepackage[colorlinks,bookmarksopen,bookmarksnumbered,citecolor=black,urlcolor=black,linkcolor=black]{hyperref}
\usepackage{flushend}

\def\volumeyear{2018}

\usepackage{graphics} 

\usepackage{subfigure}
\usepackage{epsfig} 
\usepackage{amsmath} 
\usepackage{amssymb}  
\usepackage[linesnumbered, lined, ruled]{algorithm2e}
\usepackage{epstopdf}
\usepackage{bm}
\usepackage{stfloats}
\usepackage{array}
\usepackage[T1]{fontenc}
\usepackage{booktabs}
\usepackage[table]{xcolor}
\usepackage{color, colortbl}
\usepackage{lastpage}
\usepackage{multirow}
\usepackage{todonotes}

\usepackage[normalem]{ulem}
\definecolor{Gray}{gray}{0.7}

\def\fps@figure{htp}
\def\fps@table{htp}

\newcommand{\bfig}{\begin{figure}}
\newcommand{\efig}{\end{figure}}

\newcommand{\benum}{\begin{enumerate}}
\newcommand{\eenum}{\end{enumerate}}

\newcommand{\ba}{\begin{eqnarray}}
\newcommand{\ea}{\end{eqnarray}}
\newcommand{\unit}[1]{\mbox{$\rm \,#1$}}

\runninghead{Zhang \textit{et~al.}}

\title{Adversarial Discriminative Sim-to-real Transfer of Visuo-motor Policies}

\author{Fangyi Zhang\affilnum{1,2}, J\"urgen Leitner\affilnum{1,2}, Zongyuan Ge\affilnum{1,3}, Michael Milford\affilnum{1,2}, and Peter Corke\affilnum{1,2}}

\affiliation{\affilnum{1}Australian Centre for Robotic Vision (ACRV), Brisbane, QLD, Australia\\
\affilnum{2}Queensland University of Technology (QUT), Brisbane, QLD, Australia\\
\affilnum{3}Monash University, Melbourne, VIC, Australia}

\corrauth{Fangyi Zhang, QUT Gardens Point Campus, S Block 1130-21, 2 George Street, Brisbane, QLD 4000, Australia}

\email{fangyi.zhang@roboticvision.org, fangyi.zhang@hdr.qut.edu.au}

\begin{abstract}

Various approaches have been proposed to learn visuo-motor policies for real-world robotic applications. One solution is first learning in simulation then transferring to the real world. In the transfer, most existing approaches need real-world images with labels. However, the labelling process is often expensive or even impractical in many robotic applications. In this paper, we propose an adversarial discriminative sim-to-real transfer approach to reduce the cost of labelling real data. The effectiveness of the approach is demonstrated with modular networks in a table-top object reaching task where a 7 DoF arm is controlled in velocity mode to reach a blue cuboid in clutter through visual observations. The adversarial transfer approach reduced the labelled real data requirement by 50\%. Policies can be transferred to real environments with only 93 labelled and 186 unlabelled real images. The transferred visuo-motor policies are robust to novel (not seen in training) objects in clutter and even a moving target, achieving a 97.8\% success rate and 1.8 cm control accuracy.
\end{abstract}


\begin{document}
\maketitle

\section{Introduction}
\label{sec:intro}

The advent of large datasets and sophisticated machine learning models, commonly referred to as deep learning, has in recent years created a trend away from hand-crafted solutions towards more data-driven ones. Learning techniques have shown significant improvements in robustness and performance since early work~\citep{krizhevsky2012imagenet}, particularly in the computer vision field.

Traditionally robotic vision-based reaching approaches have been based on crafted controllers that combine (heuristic) motion planners with the use of hand-crafted features to localize the target visually. Recently learning approaches to tackle this problem have been presented~\citep{zhang2015towards,zhang2017acra,zhang2017cvpr,levine2016learning,bateux2017visual,katyal2017leveraging,dlinrobotics2018}, however a consistent issue faced by most approaches is the reliance on large amounts of data to train these models. Generalization forms another challenge: many current systems are brittle when learned models are applied to robotic configurations or scenarios that differ from those used in training.
This leads to the question: 
\textit{How to better learn and transfer visuo-motor policies on robots for tasks such as reaching?}

\begin{figure}[tb!]
\begin{center}
\includegraphics[width=1\columnwidth]{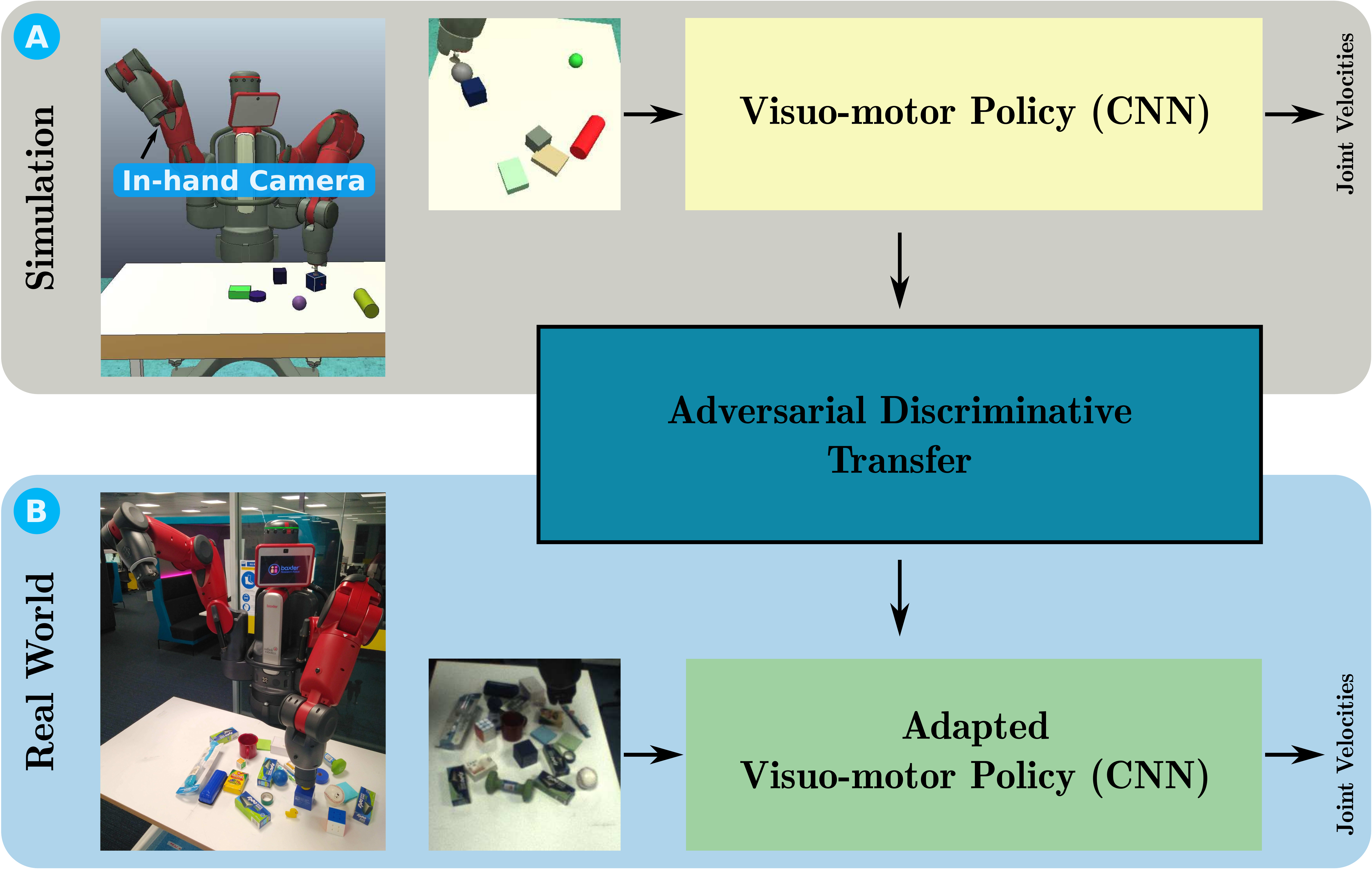}
\end{center}
\caption{A robot (Baxter) learns visuo-motor policies in simulation (Fig.~\ref{fig:intro}A) to control its left arm (7 DoF) to reach a target blue cuboid in clutter on a table. Baxter visually observes the table-top environments through the monocular camera in its right hand. An adverversarial discriminative approach (see Section~\ref{sec:methodology}) is used to transfer visuo-motor policies from simulation to the real world (Fig.~\ref{fig:intro}B). The transfer is in a semi-supervised manner which needs very few labelled real images.}
\label{fig:intro}
\end{figure}

Various approaches have been proposed to address this problem. Some works tried to directly learn from large-scale real-world datasets~\citep{levine2016learning,pinto2016supersizing}. However, collecting a large amount of real data could be expensive in robotic applications. For example, an "arm farm" with 6 to 14 physical robots was developed to collect data in parallel for learning robotic grasping~\citep{levine2016learning}. Therefore, some methods were proposed to reduce the cost of collecting a large amount of real-world data by using simulated or synthetic data~\citep{bateux2017visual,d2017bridging,tobin2017domain,james2017transferring}. Some others tried to make use of both simulated and real data for a more balanced solution~\citep{fitzgerald2015similarity,tzeng2016adapting}.
A particular approach is modular deep Q-networks for learning a planar reaching task in simulation and then transferring to real environments with a small number of labelled real-world images~\citep{zhang2017acra,zhang2017cvpr}.

In this work, we extend the modular approach~\citep{zhang2017acra} and focus on making use of both simulated and real data to learn robotic skills. In the modular deep Q-networks, labelled real images were previously used. Although the amount was small, the data labelling cost was nontrivial. In comparison, images themselves are cheap for a vision-based robotic system. Aiming for more data-efficient learning, an adversarial approach similar to GANs~\citep{goodfellow2014generative} was proposed to learn a classifier for grasping using labelled synthetic and unlabelled real data~\citep{bousmalis2018}. However, most existing works used adversarial approaches for classification tasks such as incremental adversarial domain adaptation for drivable-path segmentation~\citep{wulfmeier2017incremental}. To the best of our knowledge, there is no existing work using adversarial methods for the transfer of regression tasks.

In this paper, we propose an adversarial discriminative approach for regression transfer and investigate its effectiveness for learning visuo-motor policies from simulation to the real world. Our approach is verified with modular networks in a visually-guided table-top object reaching task for a 7 DoF robotic arm (Fig.~\ref{fig:intro}). By introducing an adversarial loss, visuo-motor policies can be successfully transferred from simulated (Fig.~\ref{fig:intro}A) to real (Fig.~\ref{fig:intro}B) environments with only 93 labelled and 186 unlabelled real images. Benefiting from the modular structure and weighted end-to-end fine-tuning, the learned visuo-motor policies achieved a reaching accuracy of 1.8\unit{cm} with only 333 trajectories (30225 state-velocity pairs collected in simulation). The learned visuo-motor policies are not only able to reach a target object in clutter with seen distractor objects, but also for the cases with novel (not seen in training) distractor objects and even when the target object is moving. 
In particular, this paper has three major contributions:
\begin{itemize}
    \item Introduction of an adversarial discriminative approach in a semi-supervised manner for more data-efficient perception transfer from simulation to the real world, achieving a comparable accuracy (2.7\unit{cm}) with 50\% fewer labelled real data and a slightly worse accuracy (3.0\unit{cm}) with 75\% fewer labelled real data (compared to supervised adaptation: 2.8\unit{cm}); 
    \item Further verification of modular neural networks~\citep{zhang2017acra} for sim-to-real transfer of visuo-motor policies in a more realistic robotic reaching task: table-top object reaching in clutter using a 7 DoF arm in velocity mode, achieving a 97.8\% success rate and 1.8\unit{cm} accuracy;
    \item Investigations on important factors in our adversarial discriminative transfer approach with comprehensive comparison experiments and detailed analyses, showing their benefits and limits for future research.
\end{itemize}
\section{Related Work}
\label{sec:related_work}

Data-driven learning approaches have become popular in computer vision and are starting to replace hand-crafted solutions in robotic applications~\citep{dlinrobotics2018}.
In particular there have been growing interest in robotic vision tasks -- robotic tasks based directly on real image data -- such as object grasping and manipulation~\citep{levine2016learning, pinto2016supersizing, lenz2015deepmpc}. An important factor in data-driven robot learning approaches is large-scale datasets, from either the real world or simulation.

\subsection{Learning from Real Datasets}
In the real world, collecting the datasets required for deep learning has been sped up by using many robots operating in parallel~\citep{levine2016learning}. With over 800,000 grasp attempts recorded, a deep network was trained to predict the success probability of a sequence of motions aiming at grasping using a 7 DoF robotic manipulator with a 2-finger gripper. Combined with a simple derivative-free optimization algorithm the grasping system achieved a success rate of 80\%. Another example of dataset collection for grasping is the approach to self-supervised grasp learning in the real world where force sensors were used to autonomously label samples~\citep{pinto2016supersizing}. After training with 50,000 real-world trials using a staged leaning method, a deep convolutional neural network (CNN) achieved a grasping success rate around 70\%.

The aforementioned results are impressive but were achieved at high cost in terms of dollars, space and time (weeks to months). To reduce the cost, Levine~et~al.\ introduced a CNN-based policy representation architecture with an added guided policy search (GPS) to learn visuo-motor policies (mapping joint angles and camera images to joint torques)~\citep{levine2016end}, which allows reduction in the number of real world training examples by providing an oracle (or expert's initial condition to start learning). Impressive results were achieved in complex tasks, such as hanging a coat hanger, inserting a block into a toy, and tightening a bottle cap.

\subsection{Learning with Simulation}

Simulation is another resource to reduce the cost of collecting real-world datasets. With domain randomization, policies learned in simulation are robust enough to be directly used on real robots with real RGB cameras observing real scenes in manipulation tasks~\citep{tobin2017domain,james2017transferring}. Recently it has also been proposed to simulate depth images to learn and then directly transfer grasping skills to real-world robotic arms~\citep{viereck2017learning}.

There are also some negative results, which show that visuo-motor policies learned in a low-fidelity simulator do not transfer directly to real robots with real cameras observing real scenes~\citep{zhang2015towards}. In fact very modest image distortions in the simulation environment (small translations, Gaussian noise and scaling of the RGB color channels) caused the performance of the system to fall dramatically. Introducing a real camera observing the game screen was even worse~\citep{tow2016robustness}. However, if adapting with a small number of real images, the visuo-motor policies learned in a low-fidelity simulator can be well transferred to real scenarios for a robotic planar reaching task~\citep{zhang2017acra,zhang2017cvpr}.

\begin{figure*}[tb!]
\begin{center}
\includegraphics[width=2\columnwidth]{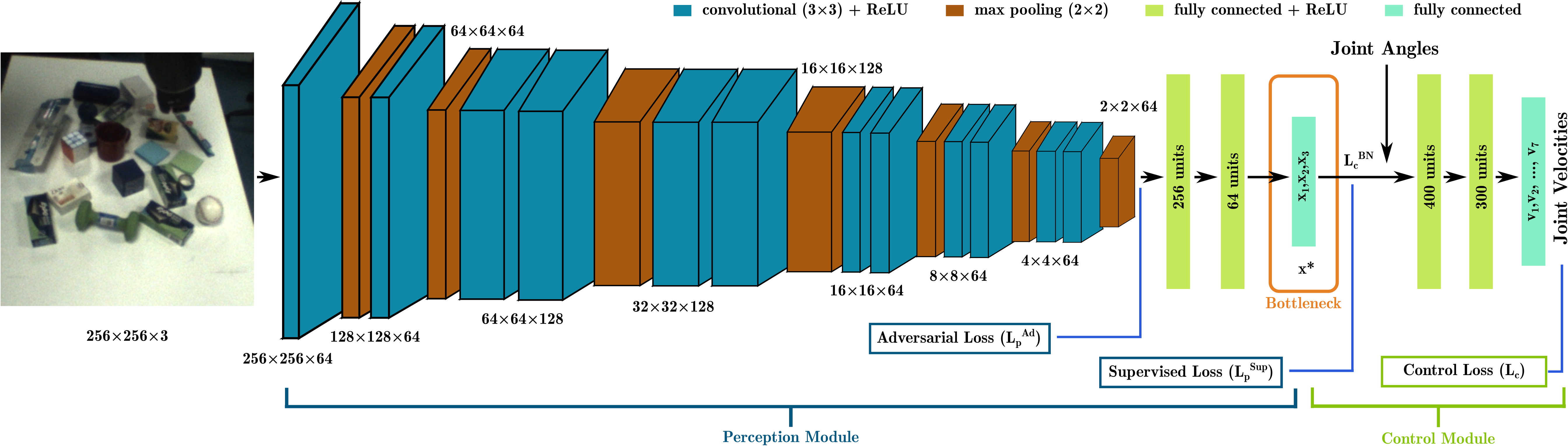}
\end{center}
\caption{The modular network consists of perception and control modules, connected by a bottleneck layer representing target object position $\mathbf{x}^*$. The perception module architecture is customized from VGG16~\citep{simonyan2014very} with its first convolutional layer initialized with weights from pre-trained VGG16. The control module consists of 3 fully connected layers, it determines joint velocities according to target position and joint angles. The perception and control modules are first trained separately, then fine-tuned in an end-to-end fashion using weighted losses (Section~\ref{sec:training_meth}).}
\label{fig:net_arch}
\end{figure*}

\subsection{Transfer Learning}

Transfer learning attempts to develop methods to transfer knowledge between different tasks (scenarios)~\citep{pan2010survey,taylor2009transfer}. To reduce the amount of data collected in the real world (expensive), transferring skills from simulation to the real world is an attractive alternative. For the case of pre-training in simulation then adapting with very few real-world samples, appropriate transfer learning approaches are required.

To reduce the number of real-world images required for learning visuo-motor policies, a method of adapting visual representations from simulated to real environments was proposed, achieving a success rate of 79.2\% in a ``hook loop'' task, with 10 times fewer real-world images~\citep{tzeng2016adapting}. Another example of vision-based policy transfer is progressive neural networks, which are proposed to improve transfer and avoid catastrophic forgetting when learning complex sequences of tasks~\citep{rusu2016progressive}. Their effectiveness has been validated on reinforcement learning tasks such as Atari and 3D maze game playing as well as simulated robotic manipulation~\citep{rusu2016sim}.

Similar to GANs~\citep{goodfellow2014generative}, adversarial approaches are also proposed for domain adaptation in classification contexts such as handwritten digit recognition~\citep{tzeng2017adversarial,luo2017label,ge2017generative}, place classification and segmentation~\citep{wulfmeier2017addressing,wulfmeier2017incremental}. Another similar approach is domain confusion, whose feasibility has been verified in object recognition~\citep{tzeng2015simultaneous} and fine-grained recoginition~\citep{gebru2017fine}. An adversarial adaptation approach was also proposed to improve the efficiency of learning a classifier to determine whether a grasp command will be successful or not~\citep{bousmalis2018}. These approaches enabled data-efficient domain adaptation for classification tasks, but we have not found any work using adversarial methods for regression tasks to the best of our knowledge.
\section{Methodology}
\label{sec:methodology}

In our previous work~\citep{zhang2017acra}, a modular structure and its training approach were proposed to transfer visuo-motor policies from simulation to the real world in a low-cost manner. The transfer was achieved by using 1418 labelled real images to fine-tune a perception module pre-trained in simulation. In this paper, we propose a semi-supervised transfer approach to reduce the required amount of labelled real images. We call this semi-supervised approach Adversarial Discriminative Transfer (ADT) which mainly benefits from the introduction of an adversarial loss~\citep{tzeng2017adversarial}.

\subsection{Modular Deep Networks}
\label{sec:modular_net}

Similar to the modular deep Q-network~\citep{zhang2017acra}, a modular network architecture (Fig.~\ref{fig:net_arch}) is proposed, which consists of perception and control modules connected by a bottleneck layer. The \textbf{bottleneck} forces the network to learn a low-dimensional representation, not unlike Auto-encoders~\citep{hinton2006reducing}. The difference is that we \textbf{explicitly equate} the bottleneck layer with the object position ($\mathbf{x}^* \in \mathbb{R}^3$ -- ignoring orientation).

With the bottleneck, the perception module learns how to estimate the object position $\mathbf{x}^*$ from a raw-pixel image $I$; the control module learns to determine the most appropriate joint velocities $\mathbf{v}$ given the object position $\mathbf{x}^*$ and joint angles $\mathbf{q}$ (defined as scene configuration $\mathbf{\Theta}=[\mathbf{x}^*,\mathbf{q}]$). The values of  $\mathbf{x}^*$ and $\mathbf{q}$ are normalized to the interval $[0,1]$.

\subsubsection{Training Method}
\label{sec:training_meth}

\paragraph{Perception}
\label{sec:perception}

The perception module is first pre-trained using labelled simulated data with a supervised loss $L_p^{Sup}$. Then it is adapted using both simulated and real data with a compound loss
\begin{equation}
\label{equ:perception_cost}
L_p= L_p^{Sup} + L_p^{Ad},
\end{equation}
where $L_p^{Ad}$ is an adversarial loss. Definitions of the loss functions will be introduced in Section~\ref{sec:meth_adt}. We call this perception training approach adversarial discriminative transfer.

\paragraph{Control}
\label{sec:ctrl}

The control module is trained using supervised learning with only simulated data 
\begin{equation}
\label{equ:control_cost}
L_c=\frac{1}{2m} \sum_{j=1}^{m} \left \| y_c(s_j) - \mathbf{v}_j \right \|^2,
\end{equation}
where $y_c(s_j)$ is the prediction of $\mathbf{v}_j$ for state $s_j$, here $s=\mathbf{\Theta}$; $m$ is the number of samples.

\paragraph{End-to-end fine-tuning using weighted losses}
\label{sec:e2e_ft}
To further improve hand-eye coordination, an end-to-end fine-tuning is conducted for the combined network (perception + control) after their separate training, using weighted control ($L_c$) and perception ($L_p$) losses. Note that $s=I$ in Eq.~\ref{equ:control_cost} for the end-to-end fine-tuning, rather than $\mathbf{\Theta}$. The control module is updated using only $L_c$, while the perception module is updated using the weighted loss 
\begin{equation}
\label{equ:endtoend_cost}
L= \beta L_p + (1-\beta) L_c^{BN},
\end{equation}
where $L_c^{BN}$ is a pseudo-loss which reflects the loss of $L_c$ in the bottleneck; $\beta \in [0,1]$ is a balancing weight. From the backpropagation algorithm~\citep{lecun-88}, we can infer that $\delta_L = \beta \delta_{L_p} + (1-\beta) \delta_{L_c^{BN}} $, where $\delta_L$ is the gradients resulting from $L$; $\delta_{L_p}$ and $\delta_{L_c^{BN}}$ are the gradients resulting respectively from $L_p$ and $L_c^{BN}$ (equivalent to that resulting from $L_c$ in the perception module).

\subsection{Adversarial Discriminative Transfer}
\label{sec:meth_adt}

\begin{figure}[tb!]
\begin{center}
\includegraphics[width=1\columnwidth]{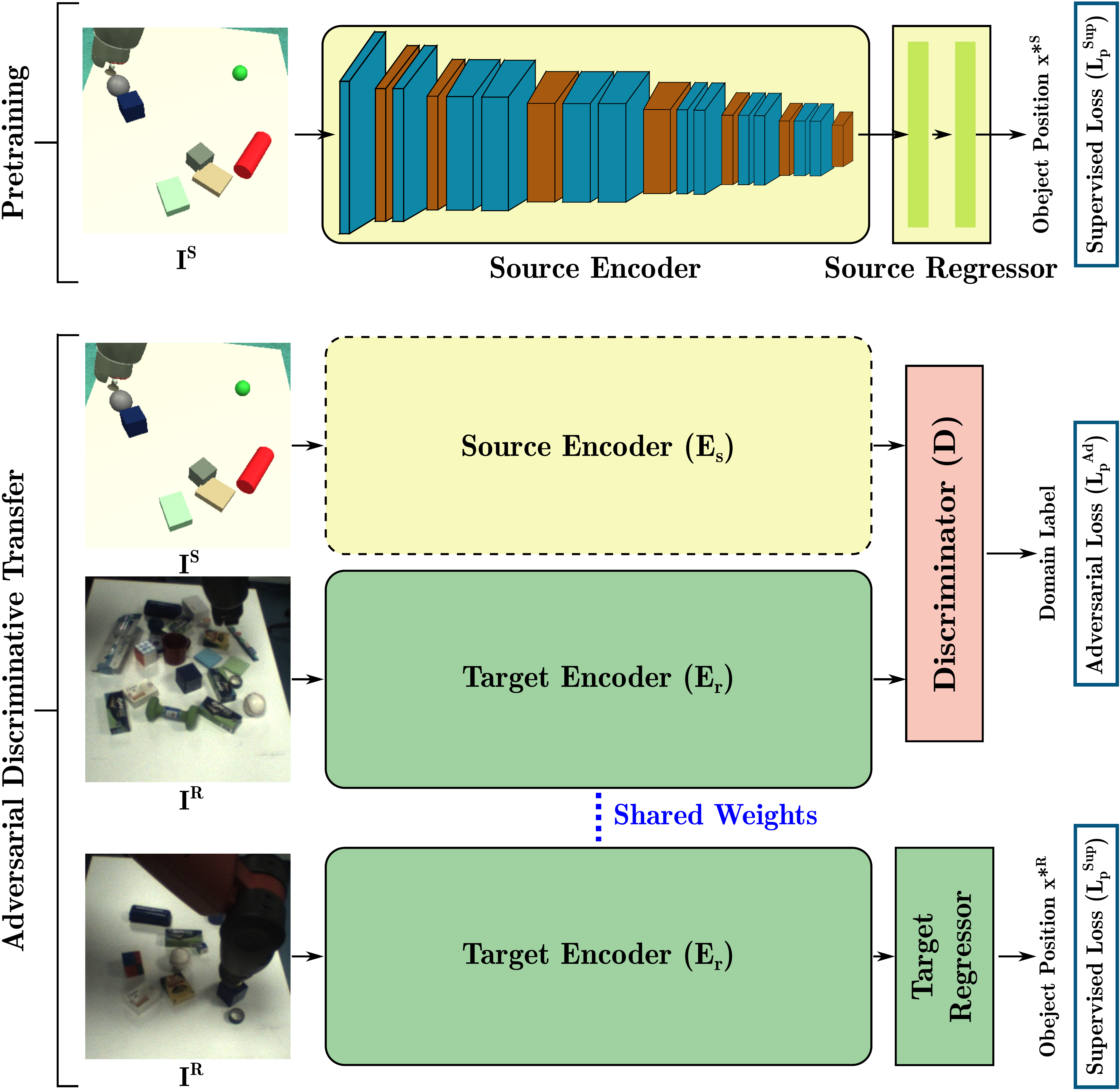}
\end{center}
\caption{In Adversarial Discriminative Transfer (ADT), the perception module is divided into two parts: encoder and regressor. The encoder includes all the convolutional layers; the regressor represents all the rest fully connected layers. We first pre-train a perception module (Source Encoder + Source Regressor) with $L_p^{Sup}$ using simulated images ($I^S$) and their target object position labels (${\mathbf{x}^*}^S$). The source encoder is then locked and used as a reference in the adversarial discriminative transfer to train a target encoder $E_r$ with $L_p^{Ad}$ using both simulated ($I^S$) and real ($I^R$) images without labels. In addition to the adversarial loss, $L_p^{Sup}$ is also used to train the target encoder and regressor with a small number of labelled real images ($I^R$ and ${\mathbf{x}^*}^R$). The target encoder and regressor are initialized with the weights in the source encoder and regressor. The discriminator consists of multiple fully connected layers.}
\label{fig:adt}
\end{figure}

Adversarial Discriminative Transfer (ADT) makes use of both adversarial and supervised losses to adapt a perception module with fewer labelled real images.
In ADT, the perception module is divided into two parts: encoder and regressor. As shown in Fig.~\ref{fig:adt}, the encoder includes all the convolutional layers in a perception module; the regressor represents all the fully connected layers of the perception module.

A perception module (Source Encoder + Source Regressor) is first pre-trained with simulated images ($I^S$) and their target object position labels (${\mathbf{x}^*}^S$), using the supervised loss
\begin{equation}
\label{equ:perception_cost}
L_p^{Sup}=\frac{1}{2m} \sum_{j=1}^{m} \left \| y_p(I_j) - {\mathbf{x}^*}_j \right \|^2,
\end{equation}
where $y_p(I_j)$ is the prediction of ${\mathbf{x}^*}_j$ for $I_j$. Here in the pre-training $I=I^S$, $\mathbf{x}^*={\mathbf{x}^*}^S$. The physical meaning of $\mathbf{x}^*$ guarantees the convenience of collecting labelled training data. 

The source encoder is then locked and used as a reference in the adversarial discriminative transfer to train a target encoder with both simulated ($I^S$) and real ($I^R$) images, but without labels, using an adversarial loss
\begin{equation}
\label{equ:adversarial_loss}
L_p^{Ad}=L_D^{Ad}+\gamma L_E^{Ad},
\end{equation}
\begin{equation}
\label{equ:D_loss}
\begin{split}
L_D^{Ad} = -\frac{1}{2m} \sum_{j=1}^{m} & \left[ \log D(E_s(I^S_j)) \right.\\
& \left. + \log (1-D(E_r(I^R_j))) \right],
\end{split}
\end{equation}
\begin{equation}
\label{equ:E_loss}
L_E^{Ad}=-\frac{1}{m} \sum_{j=1}^{m} \log D(E_r(I^R_j)),
\end{equation}
where $\gamma$ is a balancing weight; $D$ represents the discriminator; $E_s$ and $E_r$ are the source and target encoders in Fig.~\ref{fig:adt}.
With $L_D^{Ad}$, the discriminator ($D$) learns to distinguish which domain an encoded feature comes from: simulation or real world, i.e., $\arg \underset{D}{\min}\ L_D^{Ad}$. $L_E^{Ad}$ leads the target encoder ($E_r$) to be as similar as possible to the source encoder to confuse the discriminator, i.e., $\arg \underset{E_r}{\min}\ L_E^{Ad}$.

Experimental results in Section~\ref{sec:result_adt_p}) show that a single adversarial loss ($L_p^{Ad}$) is insufficient for the sim-to-real transfer of visuo-motor policies. Therefore, in addition to the adversarial loss, the supervised loss $L_p^{Sup}$ (Eq.~\ref{equ:perception_cost}) is also used in the transfer phase to train the target perception module (encoder and regressor) with a small number of labelled real images ($I^R$ and ${\mathbf{x}^*}^R$), i.e., $I=I^R$, $\mathbf{x}^*={\mathbf{x}^*}^R$.
The target perception module is initialized with the pre-trained weights from the source perception module.

\begin{figure}[tb!]
\begin{center}
\includegraphics[width=1\columnwidth]{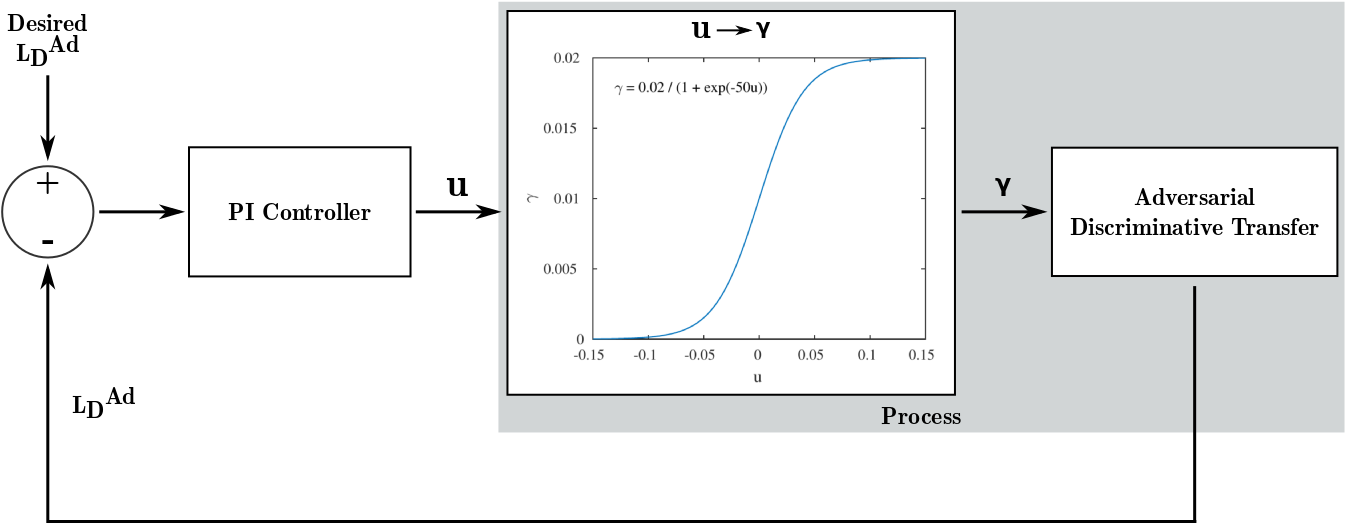}
\end{center}
\caption{A PI controller is used to control $L_D^{Ad}$ to a desired value (Desired $L_D^{Ad}$). The controller output $u$ is mapped to the balancing weight $\gamma$ through a sigmoid function. In the adversarial discriminative transfer process (Eq.~\ref{equ:adversarial_loss}), a larger $\gamma$ will result in a stronger effect of $L_E^{Ad}$ which will then more strongly prevent $L_D^{Ad}$ from being smaller or even cause a larger $L_D^{Ad}$. Similarly, a smaller $\gamma$ will help result in a smaller $L_D^{Ad}$.}
\label{fig:pid}
\end{figure}

Experimental results also show that maintaining $L_D^{Ad}$ in a certain range (0.26-0.30) helps improve the transfer performance (Section~\ref{sec:results_pid}). Therefore, a PI controller is proposed to control $L_D^{Ad}$ to a desired value by changing the balancing weight $\gamma$, as shown in Fig.~\ref{fig:pid}. In the adversarial discriminative transfer (Eq.~\ref{equ:adversarial_loss}), a larger $\gamma$ will result in a stronger effect of $L_E^{Ad}$ which will then more strongly prevent $L_D^{Ad}$ from being smaller or even cause a larger $L_D^{Ad}$. Similarly, a smaller $\gamma$ will help result in a smaller $L_D^{Ad}$.
The controller output $u$ is mapped to the balancing weight $\gamma$ through a sigmoid function $\gamma=\frac{0.02}{1+e^{-50u}}$. The sigmoid function is selected empirically according to three major concerns:
\begin{itemize}
    \item $\gamma$ cannot be too large, in order to avoid catastrophic weight forgetting;
    \item the value of $\gamma$ when $u=0$ should be able to roughly guarantee an unchanged $L_D^{Ad}$, providing a symmetric $u$-to-action-effect mapping;
    \item $\gamma$ should not be zero, since the true business of $L_E^{Ad}$ is to create a good target encoder. Although making $\gamma$ negative might better help reduce $L_D^{Ad}$, it is harmful to the true role of $L_E^{Ad}$.
\end{itemize}

Our tuned coefficients for proportional and integral gains are: $K_p=0.4$; $K_i=0.008$. To solve the integral windup problem, pre-determined bounds are used to prevent the integral term from accumulating above 0.1 or below -0.1, i.e., [-0.1,0.1].

\section{Benchmark: Robotic Reaching}

We use a canonical target reaching task as a benchmark to evaluate the effectiveness of the proposed approach. The task is defined as controlling a robot arm so that its end-effector position $\mathbf{x} \in \mathbb{R}^3$ in operational space moves to the position of a target $\mathbf{x}^* \in \mathbb{R}^3$ (object position introduced in Section~\ref{sec:modular_net}). The robot's joint configuration is represented by its joint angles $\mathbf{q} \in \mathbb{R}^n$. The two spaces are related by the forward kinematics, i.e., $\mathbf{x}=\mathcal{K}(\mathbf{q})$. The reaching controller adjusts the robot configuration in velocity mode (i.e., controls joint velocities $\mathbf{v}=\dot{\mathbf{q}} \in \mathbb{R}^n$) to minimize the error between the robot's current and target position, i.e., $\left \| \mathbf{x} - \mathbf{x}^*\right \| $. We consider a 7 DoF robotic arm (Fig.~\ref{fig:intro}), i.e., $\mathbf{q}, \mathbf{v} \in \mathbb{R}^7$, steering its end-effector position in 3D -- ignoring orientation.

\subsection{Task Setup} 
\label{sec:problem}
\begin{figure}[tb!]
\begin{center}
\includegraphics[width=1\columnwidth]{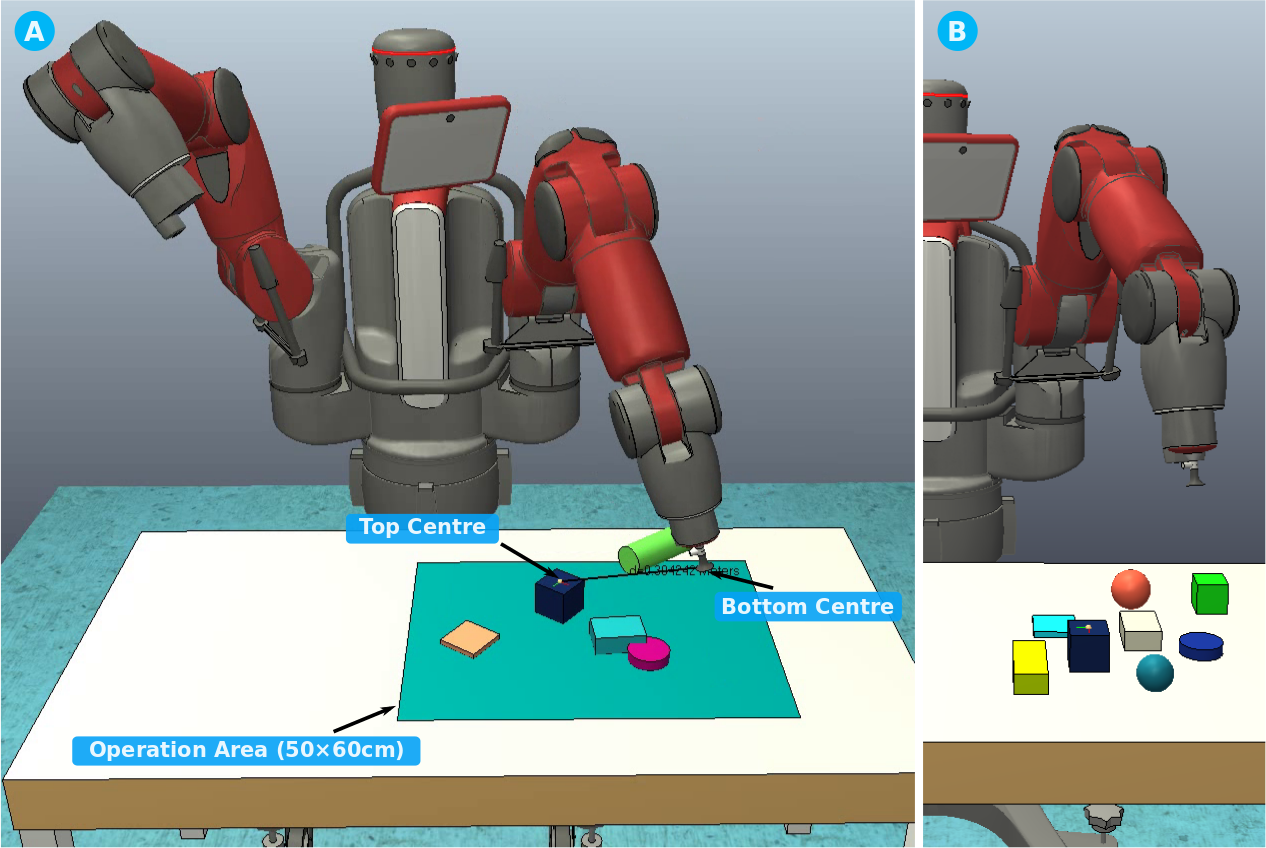}
\end{center}
\caption{A Baxter robot controls its left arm in velocity mode to reach a blue cuboid (6.5$\times$6.5$\times$6.5\unit{cm}) in clutter arbitrarily placed in the operational area. The ``Top Centre'' and ``Bottom Centre'' are the top centre of the target cuboid and the bottom center of the suction gripper. Fig.~\ref{fig:task}B shows the left arm in its reference initial configuration.}
\label{fig:task}
\end{figure}

The real-world task employs a Baxter robot's left arm (7 DoF) to reach a blue cuboid in clutter. All objects are arbitrarily placed in the operational area (50$\times$60\unit{cm}) on a table, as shown in Fig.~\ref{fig:task}A. The blue cuboid has a side length of 6.5\unit{cm}. The robot observes environments through a monocular camera in its right hand (Fig.~\ref{fig:intro}A), providing RGB images with a resolution of 256$\times$256 (cropped from 640$\times$400 images). The left arm is controlled in velocity mode. A reach is deemed successful, if the Euclidean distance between the the top centre of the target cuboid and the bottom center of the suction gripper (``Top Centre'' and ``Bottom Centre'' in Fig.~\ref{fig:task}) is smaller than 4.6\unit{cm} (half of the diagonal length of any side of the cuboid). In the task, the left arm is randomly initialized to a configuration with a normal distribution around the reference configuration shown in Fig.~\ref{fig:task}B.
The right arm is set to a constant pose, i.e., camera pose is constant with possible minor errors (Baxter joint pose accuracy: $\pm$0.10$^\circ$) in the real world.

\subsection{Network Architecture}
\label{sec:net_arch}
In this work, we used a network with the architecture shown in Fig.~\ref{fig:net_arch}. The \textbf{perception} module has an architecture customized from VGG16~\citep{simonyan2014very}. The customization mainly includes reducing the number of convolutional layers in each group (between two max pooling layers) and changing the number of feature maps in each convolutional layer for lower computational cost but without losing performance for the benchmark task.
It consists of twelve convolutional layers with 3$\times$3 filters and seven 2$\times$2 max pooling layers, followed by three fully connected layers. The twelve convolutional layers and two hidden fully connected layers use ReLU activation. Simulated or real RGB images are cropped and down-sampled to $256\times256$ as inputs to the perception module. The pixel values in images were normalized to $[-1,1]$.
The first convolutional layer is initialized with pre-trained weights for ILSVRC-2014~\citep{simonyan2014very} (observed to converge faster and achieve better performance than random initialization); other layers are randomly initialized. 

The \textbf{control} module consists of 3 fully connected layers, with 400 and 300 units in the two hidden layers (with ReLU activation) respectively. Input to the control module is the scene configuration $\mathbf{\Theta}$ (target position and joint angles), its outputs are the estimates for joint velocities $\mathbf{v}$.
This module is initialized with random weights.

The \textbf{discriminator} network consists of 3 fully connected layers with 256 units in each of its two hidden layers (also with ReLU activation). Input to the discriminator is an encoded feature vector with a dimension of 256, either from the source encoder or target encoder. The output layer has 2 units (2 classes: sim or real) with softmax activation. 
The discriminator is also randomly initialized.

\subsection{Datasets Collection}
\label{sec:datasets}

\begin{figure}[tb!]
\begin{center}
\includegraphics[width=0.95\columnwidth]{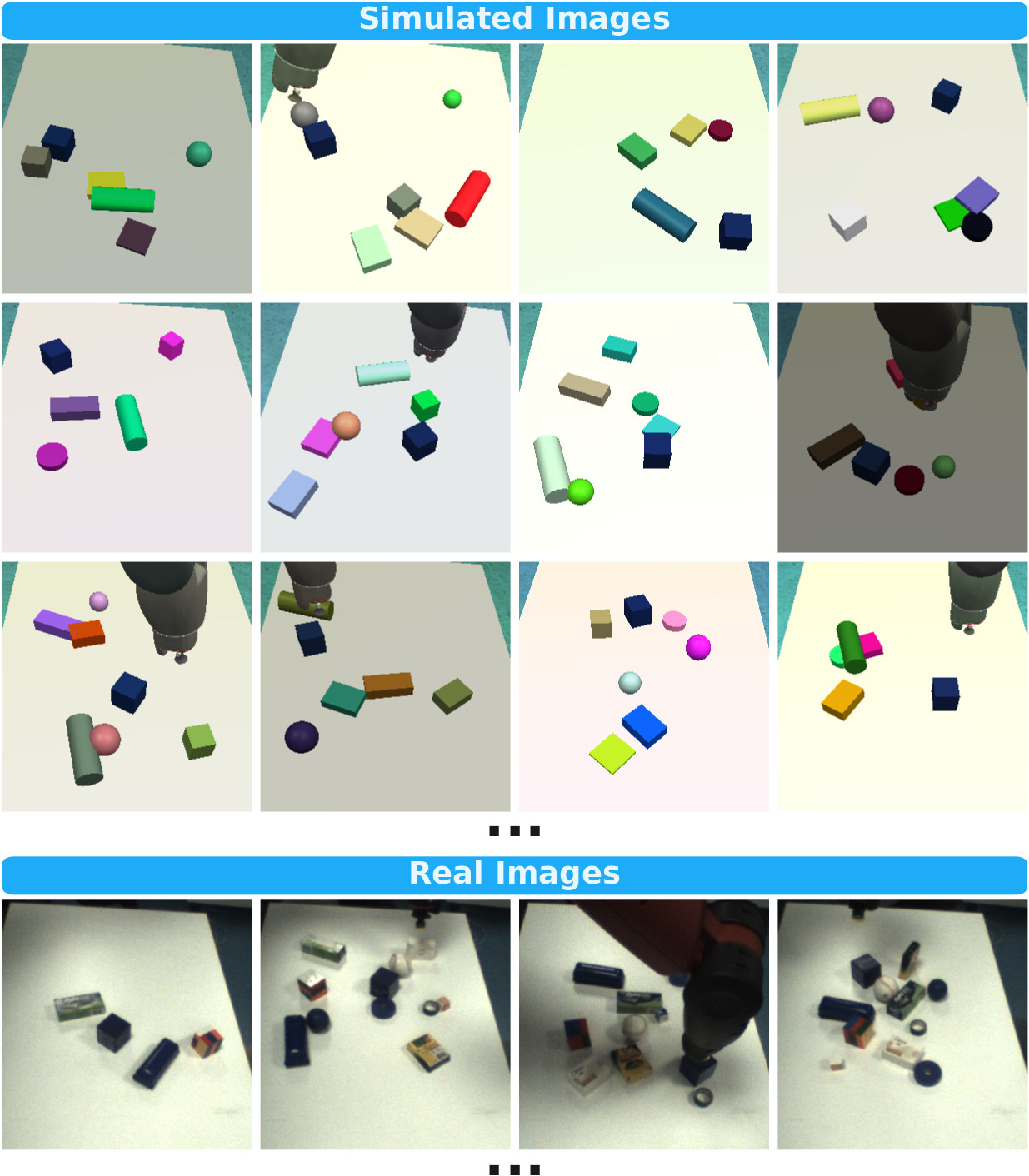}
\end{center}
\caption{Simulated and real images for training perception modules. Simulated images were collected from a V-REP simulator using domain randomization~\citep{tobin2017domain}. Real images were collected for perception adaptation on a real Baxter (Fig.~\ref{fig:intro}B).}
\label{fig:datasets}
\end{figure}

\textbf{Perception datasets} contain a number of image-position ($I$-$\mathbf{x}^*$) pairs. In this work, we label the position of the target cuboid top centre as the target position $\mathbf{x}^*$ rather than its mass-centre. 
Fig.~\ref{fig:datasets} shows some samples of the collected simulated and real images for the benchmark task. The simulated data was collected using V-REP~\citep{rohmer2013v} (a robotic simulation platform) through domain randomization~\citep{tobin2017domain} in the following aspects:
\begin{itemize}
    \item number of distractor objects in clutter: random in $[0,9]$;
    \item shape of distractor objects in clutter: random in 9 primitive shapes with different geometries (5 cuboids, 2 spheres, 2 cylinders);
    \item pose of distractor objects: random position in the operational area and random orientation about the vertical axis;
    \item color of distractor objects: random RGB values;
    \item left arm configuration: random in joint space, excluding the ones with self-collision;
    \item color of the table, floor, robot body and target cuboid: random changes based on reference colors ($\pm10\%$);
    \item camera pose: random changes of the right arm joint configuration relative to reference angles ($\pm1\%$);
    \item camera field of view (FoV): random changes based on a reference FoV ($\pm2\%$);
    \item table pose: random changes based on a reference position ($[\pm1.5\%,\pm5\%,\pm1\%]$) and a reference orientation about the vertical axis ($\pm7\%$).
\end{itemize}
All the above randomization is uniformly distributed. The reference colors, FoV and table pose were tuned manually to approximate the real scene. The reference joint angles of the right arm (i.e., camera pose) were tuned in the real world, making sure the in-hand camera can see the entire operational area. The parameters for the randomized factors based on references were manually tuned to simulate possible variations in the real scene.

The real images shown in Fig.~\ref{fig:datasets} were collected in the real world on a Baxter robot (Fig.~\ref{fig:intro}B) with random objects and left arm configurations. There are 11 real distractor objects in total. The ground-truth position of the target blue cuboid was collected by putting the end-effector bottom centre on the cuboid top centre and recording the left arm configuration (target configuration $\mathbf{q}^*$) for forward kinematics, i.e., $\mathbf{x}^*=\mathcal{K}(\mathbf{q}^*)$. The ground-truth position collected in this way is accurate enough for the benchmark task, although some errors might be caused by manually matching the end-effector with the cuboid. This ground-truth position collection method was also used in the control performance evaluation in Section~\ref{sec:result}.

More formally, we use $\mathbf{Z}_{Sup}^S(N)={\{I^S_i,{\mathbf{x}^*}^S_i\}}_{i=0}^{N}$ to represent a perception dataset of $N$ labelled simulated images. Similarly, $\mathbf{Z}_{Sup}^R(N)={\{I^R_i,{\mathbf{x}^*}^R_i\}}_{i=0}^{N}$ represents a perception dataset of $N$ labelled real images. Apart from the labelled real images, we also collected real images without labels for the adversarial discriminative transfer, represented as $\mathbf{Z}_{Ad}^R(N)={\{I^R_i\}}_{i=0}^{N}$.

In training, to increase the training data diversity, data augmentation is done on-the-fly for both simulated and real images by varying image brightness ($\pm80\%$ for simulated images and $\pm40\%$ for real images) and white balance ($\pm2.5\%$) in a post-processing manner. These augmentation parameters were empirically determined.

\begin{table*}[tb!]
    \caption{Collected Datasets}
    \label{tab:collected_datasets}
    \centering
    \renewcommand\arraystretch{1.2}
    \renewcommand\tabcolsep{20pt}
    \begin{tabular}{c|l}
    \toprule
        Simulated Perception Datasets & $\mathbf{Z}_{Sup}^S(340)$, $\mathbf{Z}_{Sup}^S(750)$, $\mathbf{Z}_{Sup}^S(3000)$\\
        \hline
        Real Perception Datasets \texttt{with} Labels & $\mathbf{Z}_{Sup}^R(48)$, $\mathbf{Z}_{Sup}^R(93)$, $\mathbf{Z}_{Sup}^R(186)$, $\mathbf{Z}_{Sup}^R(279)$\\
        \hline
        Real Perception Datasets \texttt{without} Labels & $\mathbf{Z}_{Ad}^R(48)$, $\mathbf{Z}_{Ad}^R(93)$, $\mathbf{Z}_{Ad}^R(186)$, $\mathbf{Z}_{Ad}^R(279)$\\
        \hline
        Control Datasets & $\mathbf{Z}_{c}^S(118;10677)$, $\mathbf{Z}_{c}^S(333;30225)$, $\mathbf{Z}_{c}^S(2964;269851)$\\
    \bottomrule
    \end{tabular}
\end{table*}

\textbf{Control datasets} contain a number of scene-configuration-velocity ($\mathbf{\Theta}$-$\mathbf{v}$) pairs (i.e., trajectories) as well as image-velocity ($I$-$\mathbf{v}$) and image-position ($I$-$\mathbf{x}^*$) pairs. $\mathbf{\Theta}$-$\mathbf{v}$ pairs are for training control modules separately (Section~\ref{sec:ctrl}); $I$-$\mathbf{v}$ and $I$-$\mathbf{x}^*$ pairs are for end-to-end fine-tuning to obtain $\delta_{L_c}$ and $\delta_{L_p}$ (Section~\ref{sec:e2e_ft}). 

Control datasets were purely collected in simulation using V-REP, represented as $\mathbf{Z}_{c}^S(N_T;N)={\{I^S_i,{\mathbf{x}^*}^S_i,{\mathbf{\Theta}}^S_i,{\mathbf{v}}^S_i\}}_{i=0}^{N}$ where $N_T$ indicates the number of trajectories in a dataset; $N$ is the number of samples (frames in trajectories). In dataset collection, trajectories were generated to control the left arm with a random initial configuration (excluding the ones with self-collision) to reach a target arbitrarily placed in the operational area, without considering obstacle avoidance. As introduced in Section~\ref{sec:problem}, the random initial configuration has a normal distribution around the reference configuration shown in Fig.~\ref{fig:task}B; the random targets are uniformly distributed in the operational area. When generating the trajectories, the pseudo inverse method (V-REP internal implementation) was used to calculate the desired arm configuration to reach a target, i.e., $\mathbf{q}^*=\mathcal{K}^{-1}(\mathbf{x}^*)$. Then a simple proportional controller was used to control the left arm to reach the desired configuration from its initial configuration with a control frequency of 20\unit{Hz}. In the process, the target cuboid position, joint angles and velocity commands were recorded, along with synthetic images from the camera in the right hand. Experiments (Section~\ref{sec:result_c}) show that simulated control training data is sufficient to achieve good real-world performance alone -- there is no need to collect real control datasets.

For comparison experiments in Section~\ref{sec:result}, we collected 11 perception (3 labelled simulated, 4 labelled real and 4 unlabelled real) and 3 control datasets, as listed in Table~\ref{tab:collected_datasets}. The datasets and codes will be available at \url{https://github.com/Fanleyrobot/ADT} after the paper is accepted.
\section{Experiments and Results}
\label{sec:result}

We first evaluated the performance of supervised perception adaptation as a baseline. The performance of the proposed approach was then evaluated in three aspects: adversarial discriminative perception adaptation performance, control module performance and hand-eye coordination. The important factors in ADT were also investigated with detailed comparison experiments. All the evaluations were conducted in the real world using the metrics of:
\begin{itemize}
\item \texttt{Perception Error}: the Euclidean distance between the estimated and ground-truth object positions;
\item \texttt{Control Error}: the Euclidean distance between the target cuboid top centre and  end-effector bottom centre (``Top Centre'' and ``Bottom Centre'' in Fig.~\ref{fig:task}A);
\item \texttt{Success Rate}: the percentage of successful reaching among all trials, where a reach is deemed successful if the final Euclidean distance between the target and end-effector (after the robot stops or its time is out) is smaller than 4.6\unit{cm} as defined in Section~\ref{sec:problem}.
\end{itemize}

\subsection{Supervised Perception Adaptation}
\label{sec:result_supervised_p}

Supervised adaptation is a commonly used approach in deep learning for knowledge transfer between different domains. Here, we used its performance as a baseline to compare with the proposed ADT approach. To investigate the influence of the numbers of simulated and real images on adapted perception accuracy, we evaluated 15 different perception modules. They were trained with different combinations of \textbf{labelled} images:
\begin{itemize}
    \item the number of \textbf{labelled} simulated images is from 0 to 3000;\\
    (i.e., $\mathbf{Z}_{Sup}^S(340)$, $\mathbf{Z}_{Sup}^S(750)$, and $\mathbf{Z}_{Sup}^S(3000)$)
    \item the number of \textbf{labelled} real images is from 0 to 279.\\
    (i.e., $\mathbf{Z}_{Sup}^R(93)$, $\mathbf{Z}_{Sup}^R(186)$. and $\mathbf{Z}_{Sup}^R(279)$)
\end{itemize}

\begin{figure}[tb!]
\begin{center}
\includegraphics[width=0.95\columnwidth]{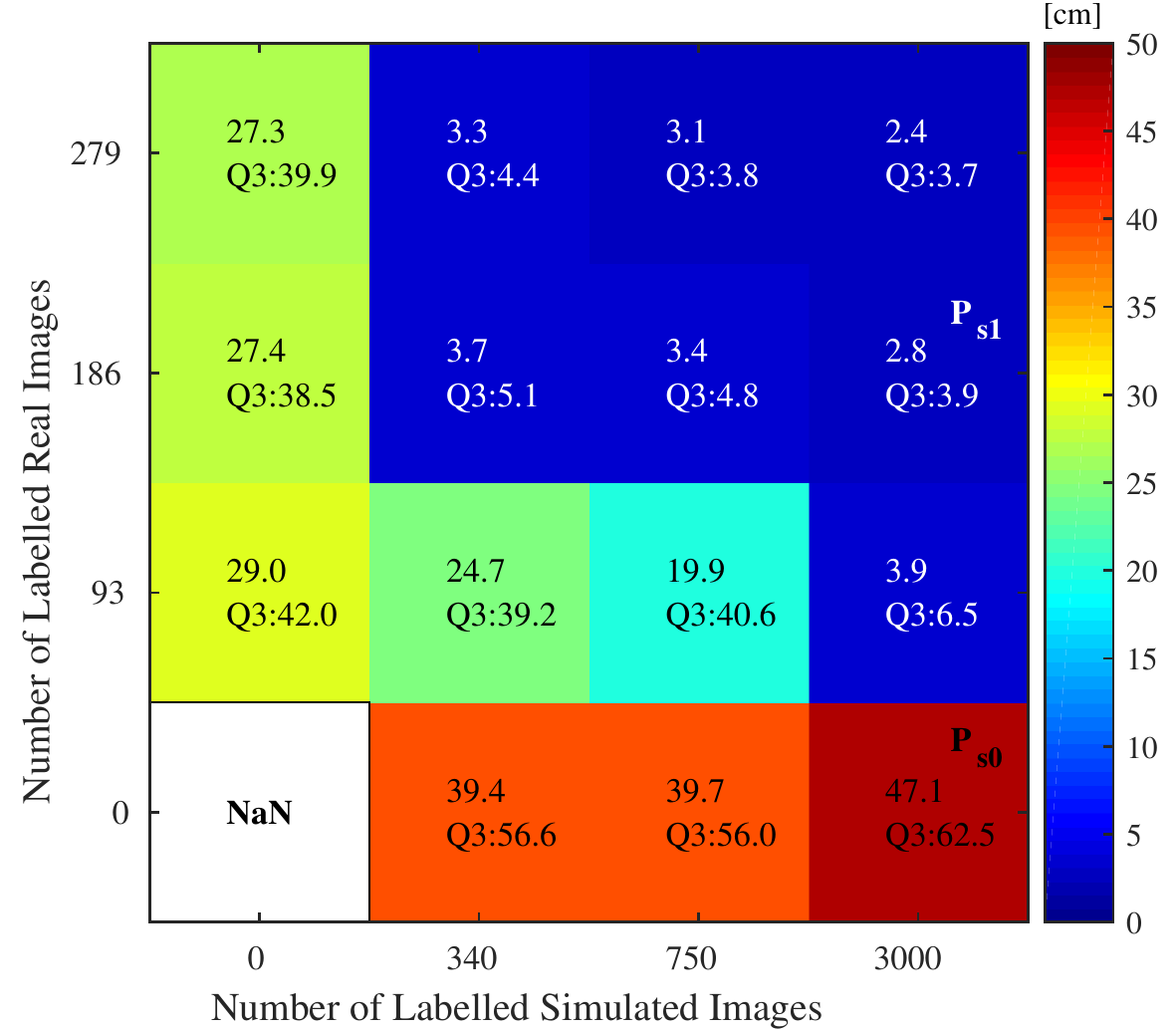}
\end{center}
\caption{Object position estimation error map for supervised adaptation. The numbers in the map show the median and third quartile (Q3) of the Euclidean distances between predicted and ground-truth positions. ``NaN'' means no result for that case.}
\label{fig:perception_map}
\end{figure}

As introduced in Section~\ref{sec:perception}, all the 15 perception modules were first trained using simulated images then adapted with real images, but only using the supervised loss $L_p^{Sup}$ without the adversarial loss. The training was from scratch, except that the first convolutional layer was initialized with weights from pre-trained VGG16~\citep{simonyan2014very}. During training, we used a mini-batch size of 32 with a learning rate of 0.01. RmsProp~\citep{tieleman2012lecture} was adopted, the same training method was used in the experiments for ADT (Section~\ref{sec:result_adt_p}), control modules (Section~\ref{sec:result_c}) and end-to-end fine-tuning (Section~\ref{sec:result_e2e}). The median and third quartile (Q3) of their perception errors for a test set are shown in Fig.~\ref{fig:perception_map}. The test set has 144 real images where the target is uniformly distributed in the operational area, with random distractor objects (those 11 distractor objects appeared in training) and left arm configurations. The test set was collected with the same setup for training set but different from those samples for training.

From Fig.~\ref{fig:perception_map}, we can see that the perception modules trained with only simulated (the bottom row) or real images (the left-most column) have very large errors. For the modules trained with both simulated and real images, increasing the number of either simulated or real images helped reduce the error. Fine-tuning (adaptation) with as few as 93 real images can make a perception module work in the real world with a median error of 3.9\unit{cm}. The module trained with 3000 simulated and 279 real images (the top-right one) achieved the smallest median error (2.4\unit{cm}). However, trading off the accuracy and the used number of real images, the module trained with 3000 simulated and 186 real images is the most balanced one, labelled as $\bold P_{\bold s1}$. It has a median error of 2.8\unit{cm} which is 17\% larger than the best one, but needs only 67\% of the real images. 
$\bold P_{\bold s0}$ which was trained with 3000 simulated samples will be used as a source perception module in the evaluation of the proposed ADT approach (Section~\ref{sec:result_adt_p}).

To study how much the on-the-fly data augmentation method (Section~\ref{sec:datasets}) can help improve the perception accuracy. We trained a perception module using 3000 simulated and 186 real images without data augmentation. 
It achieved a median error of 3.1\unit{cm} (Q3: 4.4\unit{cm}), which is 11\% larger than $\bold P_{\bold s1}$. This shows that the data augmentation did help improve the perception accuracy.

\subsection{Adversarial Discriminative Transfer}
\label{sec:result_adt_p}
In this section, we evaluated the perception modules trained by the proposed ADT approach using the same test set. 16 modules were trained using ADT to investigate how the amount of \textbf{labelled} and \textbf{unlabelled} real images influences the adaptation performance. They were adapted with different combinations of real images:
\begin{itemize}
    \item the number of \textbf{labelled} real images from 0 to 186;\\
    (i.e., $\mathbf{Z}_{Sup}^R(48)$, $\mathbf{Z}_{Sup}^R(93)$, and $\mathbf{Z}_{Sup}^R(186)$)
    \item the number of \textbf{unlabelled} real images from 0 to 279.\\
    (i.e., $\mathbf{Z}_{Ad}^R(48)$, $\mathbf{Z}_{Ad}^R(93)$, $\mathbf{Z}_{Ad}^R(186)$, and $\mathbf{Z}_{Ad}^R(279)$)
\end{itemize}

All the 16 perception modules were adapted using the adversarial loss (Eq.~\ref{equ:adversarial_loss}) from the same module $\bold P_{\bold s0}$ which was pre-trained with 3000 simulated images in Section~\ref{sec:result_supervised_p} (equivalent to the pre-training phase of ADT). 
The target encoders and regressors of the 16 perception modules were initialized with the weights of $\bold P_{\bold s0}$. The encoder part of $\bold P_{\bold s0}$ also worked as the reference source encoder in the adversarial discriminative transfer.

In the transfer phase, we used a constant learning rate of 0.001 and a mini-batch size of 32. In particular, 32 simulated (from $\mathbf{Z}_{Sup}^S(3000)$) and 32 \textbf{unlabelled} real images (from $\mathbf{Z}_{Ad}^R(N)$) were used to calculate $L_D^{Ad}$ in each transfer step; and the same 32 \textbf{unlabelled} real images were also used to calculate $L_E^{Ad}$; then another 32 \textbf{labelled} real images (from $\mathbf{Z}_{Sup}^R(N)$) were used to calculate $L_p^{Sup}$. The desired discriminative loss $L_D^{Ad}$ was set to 0.28. The other hyper-parameters are the same with that in Section~\ref{sec:result_supervised_p}.

\begin{figure}[tb!]
\begin{center}
\includegraphics[width=0.95\columnwidth]{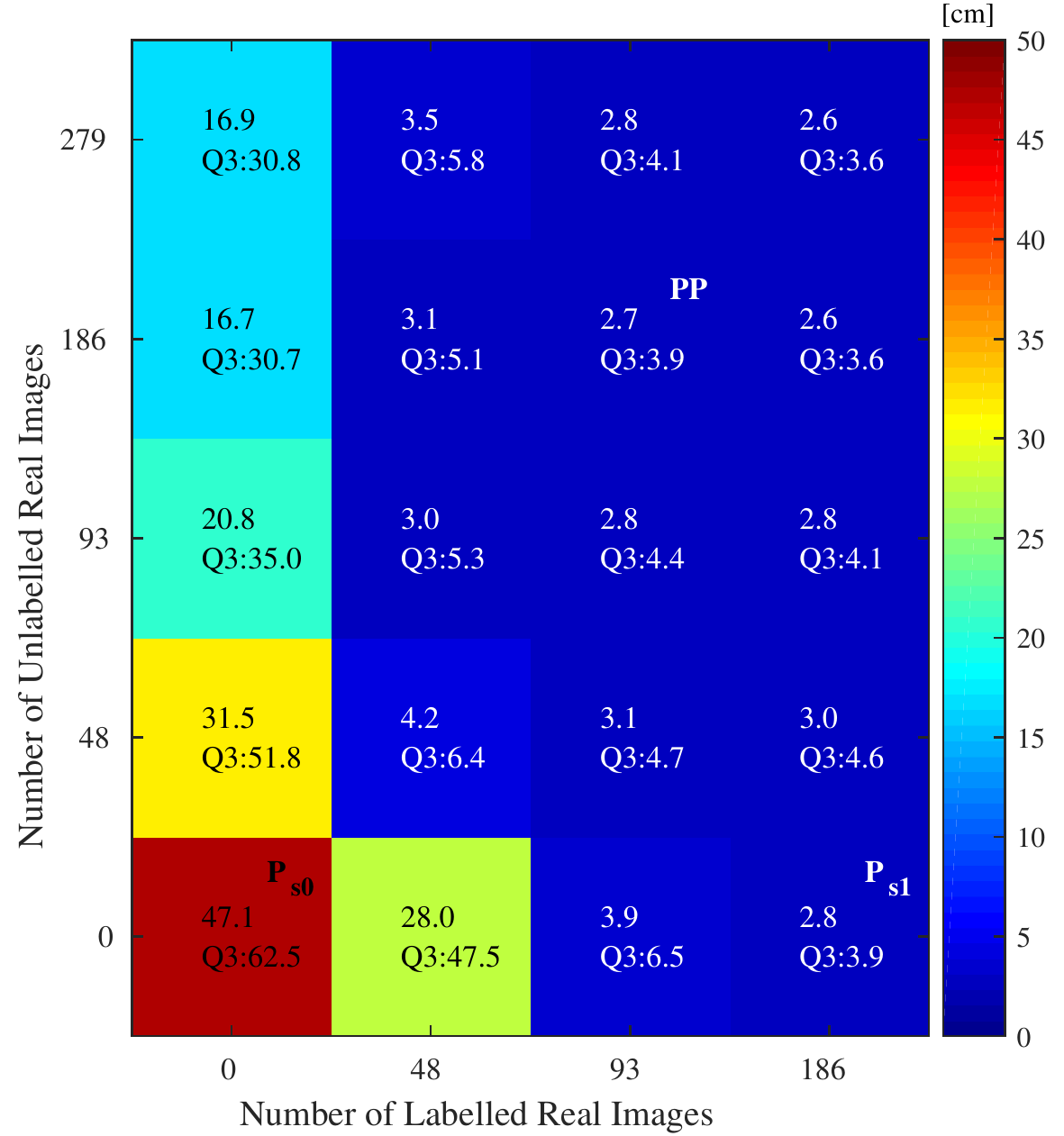}
\end{center}
\caption{Object position estimation error map for the ADT approach. Note: the x-axis shows the number of \textbf{labelled} real images used; the y-axis shows that of \textbf{unlabelled} real images.}
\label{fig:adt_map}
\end{figure}

Fig.~\ref{fig:adt_map} shows the performance of the perception modules adapted with different numbers of unlabelled and labelled real images. The bottom row shows the results for the modules adapted without unlabelled real images, i.e., supervised adaptation (three of them have appeared in Fig.~\ref{fig:perception_map}, except the one adapted with 48 labelled real images). The results for the cases without labelled real images (i.e., unsupervised adaptation, $L_p=L_p^{Ad}$) are shown in the left-most column, from which we can observe that modules adapted with more unlabelled real images have smaller errors, but marginal improvement after more than 186 images. The poor accuracy ($\geqslant 16.7$\unit{cm}) of perception modules adapted without labelled real data indicates that a single adversarial loss ($L_p^{Ad}$) is insufficient for the sim-to-real transfer of visuo-motor policies.

The other results are for the cases with both labelled and unlabelled real images (i.e., semi-supervised adaptation). We can see that the modules adapted with more labelled images have smaller errors, but the improvement is non-obvious after more than 93 labelled samples. Similarly, more unlabelled real images also resulted in smaller errors. However, performance became worse if the number of unlabelled images was more than two times the number of labelled samples (e.g., the modules adapted with 48 labelled and more than 93 unlabelled real images, as well as the module adapted with 93 labelled and 279 unlabelled samples) or fewer than half the number (e.g., the modules adapted with 186 labelled and 48 unlabelled real images). This might be because large differences between unlabelled and labelled data make their distributions differ a lot, which then result in worse adaptation. More investigation is necessary in the future to make better use of unlabelled real data, enabling performance improvement for the cases with two times more unlabelled data than labelled ones.

The best performance was achieved by the modules adapted with 186 labelled and 186 or 279 unlabelled real images. However, trading off the accuracy and the number of labelled real images (expensive), the module adapted with 93 labelled and 186 unlabelled real images is the best one, labelled as \textbf{PP}. It has a slightly larger error than the best ones, but needs 50\% fewer labelled real images.

By comparing the bottom row (supervised adaptation) with the other rows (ADT), we can see that the benefit of the adversarial loss was significant, particularly for the cases with very few labelled samples, e.g., the modules adapted with 48 labelled real images (more than 85\% improvement, perception errors reduced from 28.0\unit{cm} to less than 4.2\unit{cm}). In contrast, the benefit of the adversarial loss was trivial when adapting with 186 labelled samples.

\subsection{Important Factors in ADT}
To further investigate the effectiveness and robustness of the proposed ADT approach, we conducted some comparison experiments in four different aspects:
\begin{itemize}
    \item how robust is ADT to different random seeds in training?
    \item how effective is the PI controller?
    \item how does the desired discriminative loss for the PI controller influence the adaptation performance?
    \item how does the capacity of a discriminator network influence the adaptation performance?
\end{itemize}
In these comparison experiments, all perception modules were trained using the same conditions for \textbf{PP}, i.e., 3000 labelled simulated images ($\mathbf{Z}_{Sup}^S(3000)$), 93 labelled ($\mathbf{Z}_{Sup}^R(93)$) and 186 unlabelled ($\mathbf{Z}_{Ad}^R(186)$) real images. The training hyper-parameters other than the comparing one were the same with that for \textbf{PP}. Performances were evaluated using the same test set which was used in Section~\ref{sec:result_supervised_p} and Section~\ref{sec:result_adt_p}.

\subsubsection{Robustness to different random seeds}

\begin{figure}[tb!]
\begin{center}
\includegraphics[width=0.95\columnwidth]{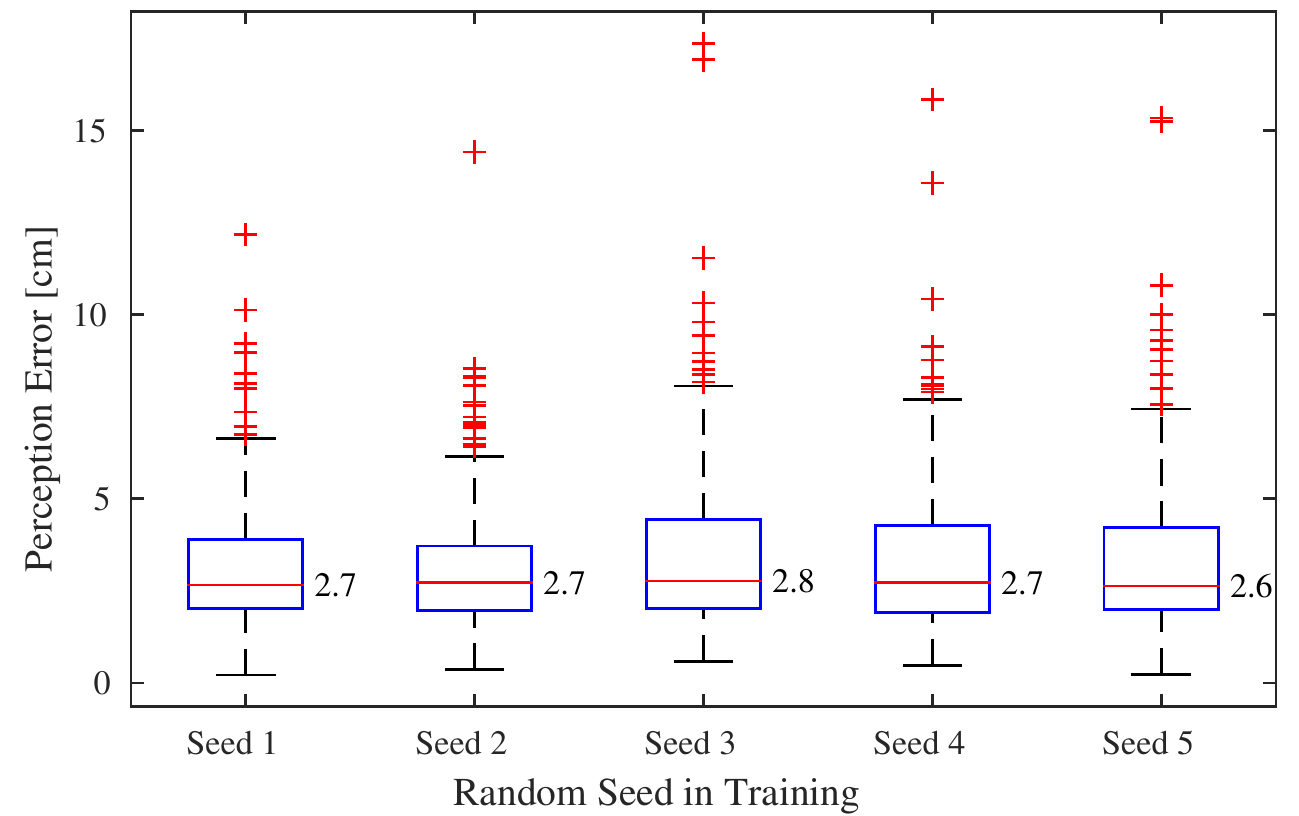}
\end{center}
\caption{The box-plots of the Euclidean distances between predicted and ground-truth positions for perception modules adapted using the ADT approach with different random seeds. The crosses represent outliers, the numbers show the medians. The outliers are the ones $\geqslant$Q3+w(Q3-Q1) or $\leqslant$Q1-w(Q3-Q1), where Q1 and Q3 are the first and third quartiles; w=1.5.}
\label{fig:boxplot_seed}
\end{figure}

To evaluate the robustness of the proposed ADT approach, 5 perception modules were trained using different random seeds, i.e., Seed 1 to 5. Seed 1 is the one used for \textbf{PP}. Fig.~\ref{fig:boxplot_seed} shows their estimation errors in the box-plot form, from which we can see that their median errors were between 2.6\unit{cm} and 2.8\unit{cm}, with a bit different distributions.
These results indicate that the ADT approach is robust to different random seeds in training.

\subsubsection{Effectiveness of the PI controller}

To see the benefit of the PI controller, a module was adapted using the adversarial loss but without the PI controller, i.e., $\gamma=1$ in Eq.~\ref{equ:adversarial_loss}. In addition, we also compared the adversarial loss to a confusion loss~\citep{tzeng2015simultaneous} where Eq.~\ref{equ:E_loss} was replaced by 
\begin{equation}
\begin{split}
  L_E^{Ad}= -\frac{1}{2m} \sum_{j=1}^{m} & \left[ \frac{1}{2}\log D(E_s(I^S_j)) \right.\\
   & \left. + \frac{1}{2}\log (1-D(E_s(I^S_j))) \right.\\
  & \left. + \frac{1}{2}\log D(E_r(I^R_j)) \right.\\
  & \left. + \frac{1}{2}\log (1-D(E_r(I^R_j)))\right ],
\end{split}
\end{equation}
of which the weights in source and target encoders were shared, i.e., $E_s=E_r$.

\begin{figure}[tb!]
\begin{center}
\includegraphics[width=0.95\columnwidth]{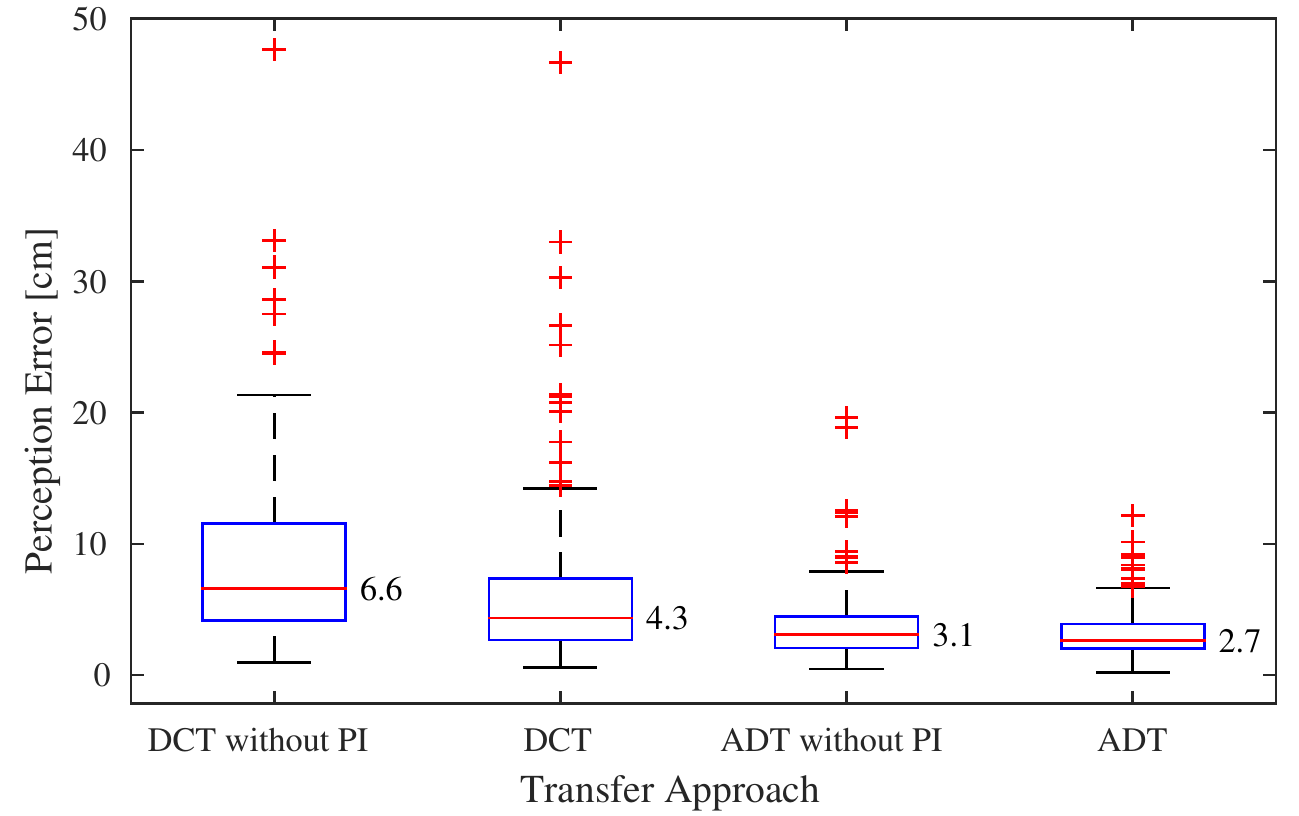}
\end{center}
\caption{The box-plots for perception modules adapted using different approaches: ADT, ADT without PI, DCT and DCT without PI.}
\label{fig:boxplot_pid}
\end{figure}

Fig.~\ref{fig:boxplot_pid} compares the results, from which we can see that the domain confusion approach (DCT) has much larger errors than ADT, either with or without the PI controller. The approaches with the PI controller achieved better performances than the ones without. In particular, ADT has a 13\% smaller median perception error than ADT without PI; DCT's median error is 35\% smaller than that of DCT without PI. These results show that the PI controller did help improve the adaptation no matter using an adversarial loss or domain confusion loss. 
The adversarial loss worked better than the domain confusion loss for our case.

\subsubsection{Appropriate desired $L_D^{Ad}$ for the PI controller}
\label{sec:results_pid}
\begin{figure}[tb!]
\begin{center}
\includegraphics[width=0.95\columnwidth]{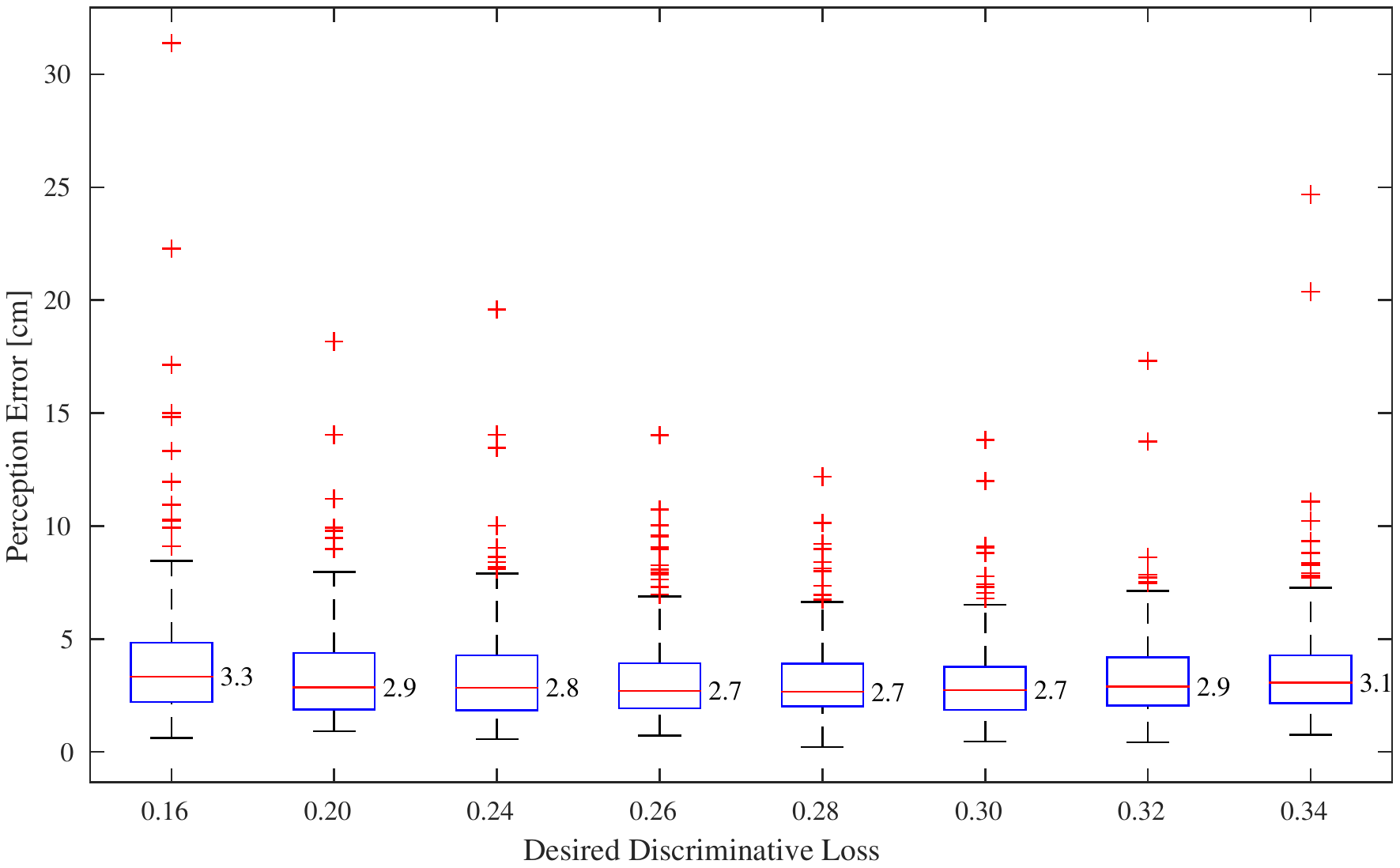}
\end{center}
\caption{The box-plots for perception modules adapted using the ADT approach with different desired discriminative losses for the PI controller.}
\label{fig:boxplot_goal}
\end{figure}

To investigate how the desired discriminative loss $L_D^{Ad}$ for the PI controller influences the adaptation performance, 8 perception modules were adapted with different desired $L_D^{Ad}$. Their estimation errors are shown in Fig.~\ref{fig:boxplot_goal}. We can see that the modules with goals between 0.26 and 0.30 have very similar performances. The others outside the interval have larger errors: the smaller or larger the desired loss is, the larger the perception error is. This shows that too large or small desired $L_D^{Ad}$ could cause worse adaptation, while setting the desired loss to a certain range (0.26-0.30) helps achieve good perception adaptation.

\subsubsection{Appropriate discriminator networks}

\begin{figure}[tb!]
\begin{center}
\includegraphics[width=0.95\columnwidth]{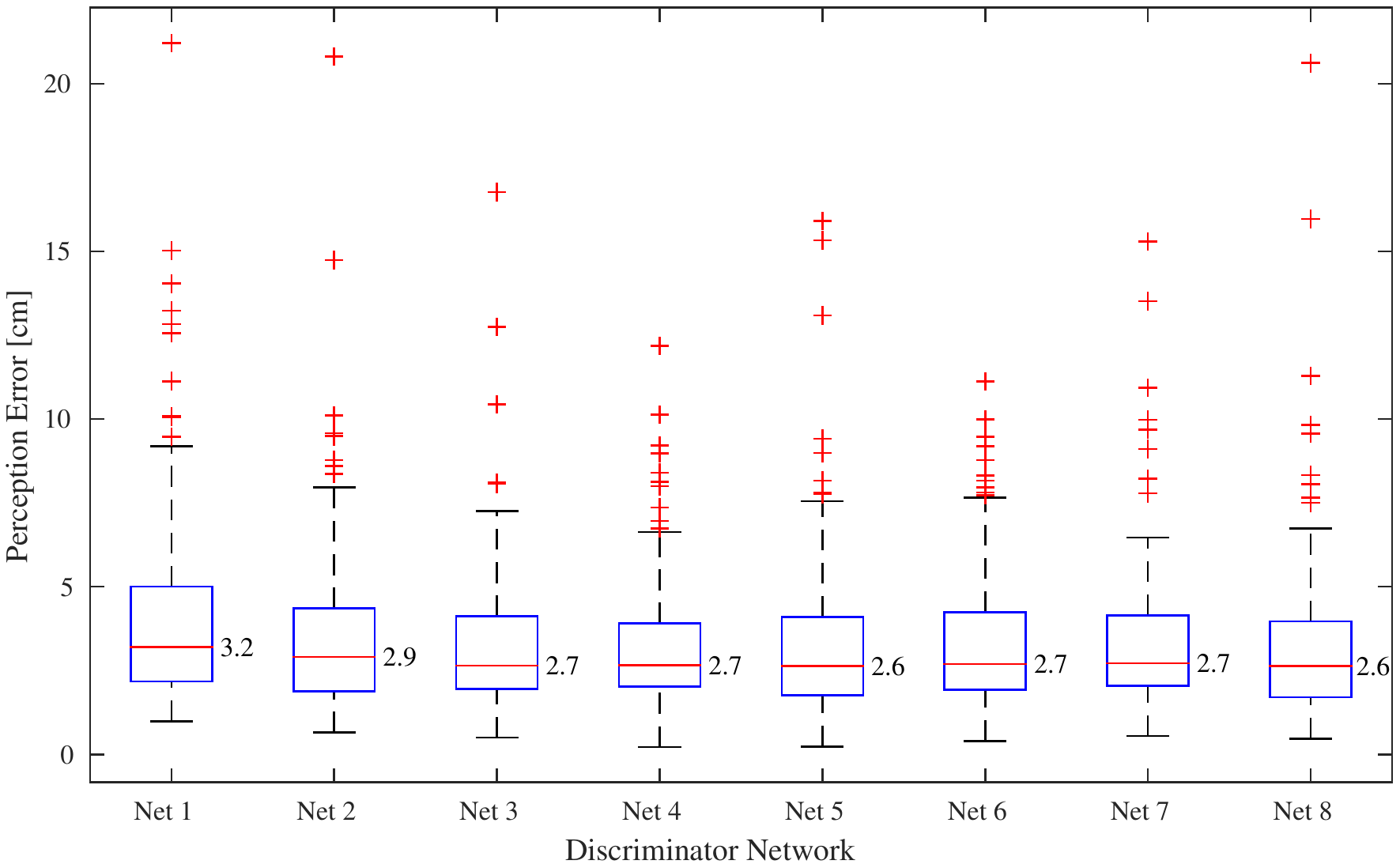}
\end{center}
\caption{The box-plots for perception modules adapted using the ADT approach with different discriminator network architectures.}
\label{fig:baxplot_dnet}
\end{figure}

To study how discriminator network architecture could influence the adaptation performance, we adapted 8 perception modules with different discriminator networks below: (numbers in brackets represent the number of units in each layer)
\begin{itemize}
    \item Net 1: 2 hidden layers, (32,32);
    \item Net 2: 2 hidden layers, (64,64);
    \item Net 3: 2 hidden layers, (128,128);
    \item Net 4: 2 hidden layers with units (256,256), the one used in other experiments;
    \item Net 5: 2 hidden layers, (512,512);
    \item Net 6: 3 hidden layers, (256,256,256);
    \item Net 7: 4 hidden layers, (256,256,256,256);
    \item Net 8: 5 hidden layers, (256,256,256,256,256).
\end{itemize}

Fig.~\ref{fig:baxplot_dnet} shows the errors of the 8 perception modules. We can see that Net 3-8 have similar perception errors with a median error of either 2.6\unit{cm} or 2.7\unit{cm}. Net 1 and Net 2 have larger errors, among which Net 1 is the worst. These results indicate that the discriminator network architecture plays an important role in the ADT approach. A discriminator with too few units in hidden layers (Net 1 and Net 2) has insufficient capacity to well distinguish the differences between simulated and real domains, therefore cannot provide enough guidance for a target encoder to be as similar as possible to a source encoder. 
For our case, a network wider or deeper than Net 3 (including Net 3) is sufficient, and makes no big difference by further widening or deepening it.

\subsection{Control Module Performance}
\label{sec:result_c}

\begin{figure}[tb!]
\begin{center}
\includegraphics[width=0.95\columnwidth]{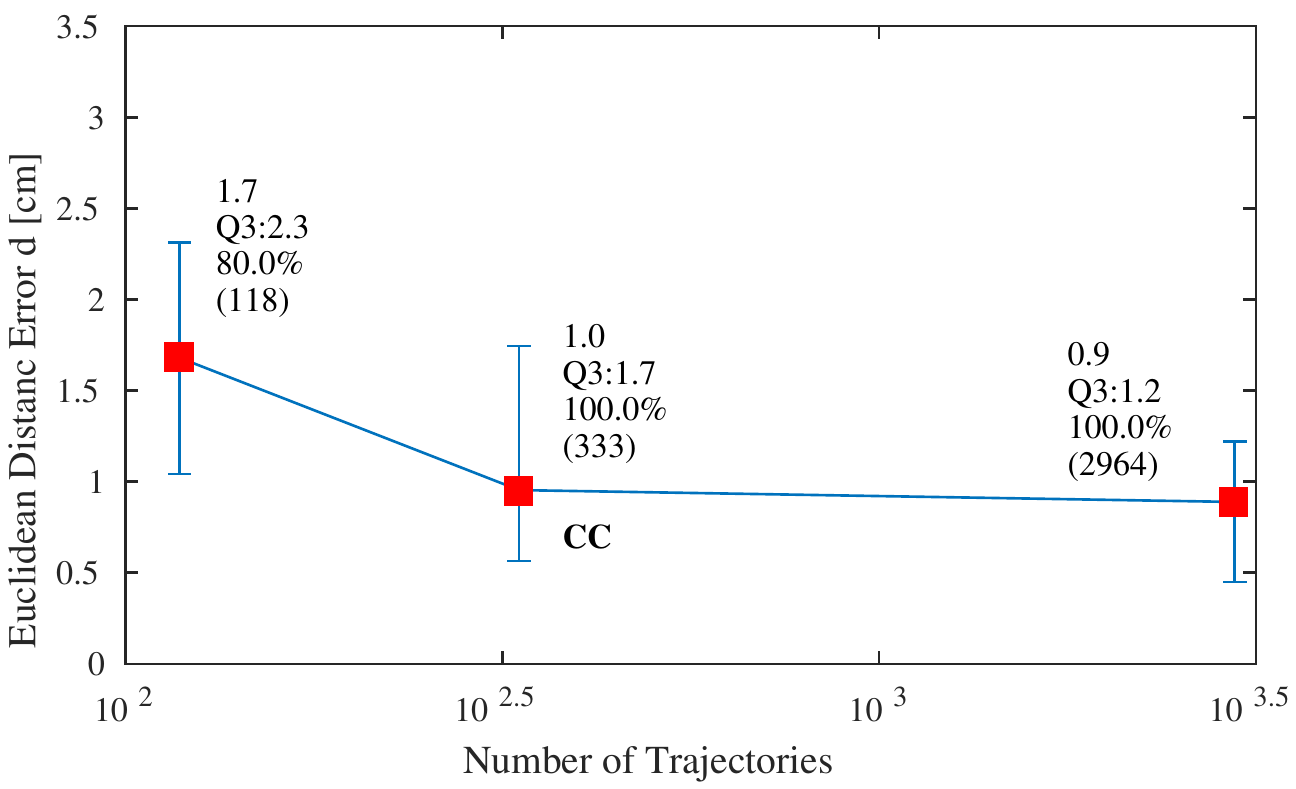}
\end{center}
\caption{Control performance curve which shows the median (red square), first quartile (Q1, lower bar) and third quartile (Q3, upper bar) of the Euclidean distances between the target and end-effector. Three control modules were evaluated in 45 trials. They were trained with different numbers of trajectories (the numbers in brackets). Their success rates are also listed.}
\label{fig:ctrl_curve}
\end{figure}

To investigate how many trajectories are sufficient for training a control module, we evaluated three control modules trained with different control datasets which have varying numbers of trajectories: 118, 333, and 2964 (i.e., $\mathbf{Z}_{c}^S(118;10677)$, $\mathbf{Z}_{c}^S(333;30225)$, and $\mathbf{Z}_{c}^S(2964;269851)$). As introduced in Section~\ref{sec:datasets}, the trajectories in each dataset were all collected in simulation (therefore cheap) for targets uniformly distributed in the operational area. In training, we used a mini-batch size of 64 and a learning rate decreasing from 0.01 to 0.001 with respect to training steps. The metrics of control error and success rate were used in the evaluation. Their performance in 45 real-world reaching trials are shown in Fig.~\ref{fig:ctrl_curve}. The 45 trials were for 15 targets (3 trials for each target) uniformly distributed in the operational area, with random initial left arm configurations (normally distributed around the reference configuration in Fig.~\ref{fig:task}B).

From Fig.~\ref{fig:ctrl_curve}, we can see that a control module trained with more trajectories is able to achieve a better control performance in terms of both control error and success rate. The control module trained with 118 trajectories has a success rate of 80\%; the other two are 100\%. It also has a much larger control error than the other two. This indicates that 118 trajectories are too few to get a good control module. The module trained with 2964 trajectories achieved a slightly smaller control error (0.9\unit{cm}, Q3:1.2\unit{cm}) than the one (1.0\unit{cm}, Q3:1.7\unit{cm}) trained with much fewer trajectories (333). This shows that 333 trajectories are sufficient to get a reasonably good control module. Trading off the performance and number of trajectories, we pick the control module trained with 333 trajectories to compose the network for end-to-end reaching in Section~\ref{sec:result_e2e}, labelled as \textbf{CC}.

\begin{table*}[tb!]
\caption{End-to-end Fine-tuning Settings}
\label{tab:e2e_experiments_settings}
\centering	
\renewcommand\arraystretch{1.2}
\begin{tabular}{m{9.4em} | m{6.4em} | m{30em}}	
\toprule 

Fine-tuning Case & Datasets & Detailed Settings\\
\hline
Naive End-to-end Fine-tuning for \textbf{EE1} & $\mathbf{Z}_{c}^S(333;30225)$ $\mathbf{Z}_{Sup}^R(186)$ & Fine-tuned \textbf{EE0} using $L_c$ (Eq.~\ref{equ:control_cost}) in an end-to-end fashion with $s=I$. A learning rate of 0.01 and a mini-batch size of 64 were used. The 186 labelled real images (from $\mathbf{Z}_{Sup}^R(186)$) for $\bold P_{\bold s1}$ were used here with velocity labels obtained using the same method for control datasets in Section~\ref{sec:datasets}. Similar to the training of $\bold P_{\bold s1}$, 87.5\% samples in a mini-batch were real ones; the simulated ones were from $\mathbf{Z}_{c}^S(333;30225)$.\\
\hline
Weighted End-to-end Fine-tuning \texttt{without} $L_p^{Ad}$ for \textbf{EE2} & $\mathbf{Z}_{c}^S(333;30225)$ $\mathbf{Z}_{Sup}^R(186)$ & End-to-end fine-tuned \textbf{EE0} using the weighted loss $L$ (Eq.~\ref{equ:endtoend_cost}) with $L_p=L_p^{Sup}$, $\beta=0.9$. A learning rate of 0.01 was used with a mini-batch size of 8 and 64 for $L_c$ and $L_p$ respectively. In each fine-tuning step, 8 random image-velocity pairs from $\mathbf{Z}_{c}^S(333;30225)$ were used to obtain $\delta_{L_c}$; its image-position pairs were used with \textbf{labelled} real images (from $\mathbf{Z}_{Sup}^R(186)$) to obtain $\delta_{L_p^{Sup}}$. Similar to the training of $\bold P_{\bold s1}$, 87.5\% samples were real ones in a mini-batch for $L_p^{Sup}$, i.e., 56 real and 8 simulated samples.\\
\hline
Weighted End-to-end Fine-tuning \texttt{with} $L_p^{Ad}$ for \textbf{EE4} &$\mathbf{Z}_{c}^S(333;30225)$ $\mathbf{Z}_{Sup}^S(3000)$ $\mathbf{Z}_{Sup}^R(93)$ $\mathbf{Z}_{Ad}^R(186)$ & End-to-end fine-tuned \textbf{EE3} using the weighted loss $L$ (Eq.~\ref{equ:endtoend_cost}) with $L_p=L_p^{Sup}+L_p^{Ad}$ (Eq.~\ref{equ:perception_cost}), $\beta=0.9$. A learning rate of 0.001 was used with a mini-batch size of 16 and 32 for $L_c$ and $L_p^{Sup}$ respectively. In each fine-tuning step, 16 image-velocity pairs from $\mathbf{Z}_{c}^S(333;30225)$ were used to obtain $\delta_{L_c}$; its image-position pairs were used with \textbf{labelled} real images (from $\mathbf{Z}_{Sup}^R(93)$) to obtain $\delta_{L_p^{Sup}}$. 
In a mini-batch for $L_p^{Sup}$, 50\% samples were real ones, i.e., 16 \textbf{labelled} real and 16 \textbf{labelled} simulated images.
The adversarial loss $L_p^{Ad}$ (Eq.~\ref{equ:adversarial_loss}) was calculated in a more complex way.
In particular, 32 simulated images (the same 16 samples from $\mathbf{Z}_{c}^S(333;30225)$ and 16 more from $\mathbf{Z}_{Sup}^S(3000)$) and 16 \textbf{unlabelled} real images (from $\mathbf{Z}_{Ad}^R(186)$) were used to calculate $L_D^{Ad}$; and the same 16 \textbf{unlabelled} real images were used to calculate $L_E^{Ad}$.\\

\bottomrule 
\end{tabular}
\end{table*}

As a comparison, we also evaluated the pseudo-inverse method (which was used to collect trajectory samples) in the real world, using joint angles and target position (not images) as inputs. It has a median control error of 0.2\unit{cm} (Q3: 0.5\unit{cm}), which is smaller than the three trained control modules. However, the control error of \textbf{CC} is small enough for our experiments, since it is much smaller than the perception error of $\bold P_{\bold s1}$ and \textbf{PP} (i.e., the control performance will not be the end-to-end performance bottleneck). Our future work will try to use reinforcement learning to further improve the control performance, but we focus on policy transfer in this paper.

\subsection{Hand-eye Coordination}
\label{sec:result_e2e}

To further improve hand-eye coordination, we proposed an end-to-end fine-tuning approach using weighted losses. To evaluate the effectiveness of the approach, we compare five combined networks and a baseline:
\begin{itemize}
    \item \textbf{Baseline}: composed by $\bold P_{\bold s1}$ and the pseudo-inverse method used to collect trajectory samples;
    \item \textbf{EE0}: composed by $\bold P_{\bold s1}$ and \textbf{CC}, directly connected after separate training without end-to-end fine-tuning;
    \item \textbf{EE1}: \textbf{EE0} end-to-end fine-tuned naively, only using the control loss $L_c$;
    \item \textbf{EE2}: \textbf{EE0} fine-tuned using the proposed approach with weighted losses, \texttt{without} $L_p^{Ad}$ (i.e., $L_p=L_p^{Sup}$);
    \item \textbf{EE3}: composed by \textbf{PP} and \textbf{CC}, directly connected after separate training without end-to-end fine-tuning;
    \item \textbf{EE4}: \textbf{EE3} fine-tuned using the proposed approach with weighted losses, \texttt{with} $L_p^{Ad}$ (i.e., $L_p=L_p^{Sup}+L_p^{Ad}$).
\end{itemize}

The detailed end-to-end fine-tuning settings for \textbf{EE1}, \textbf{EE2} and \textbf{EE4} are listed in Table~\ref{tab:e2e_experiments_settings}. In the naive fine-tuning for \textbf{EE1}, extra velocity labels were collected for the real images in $\mathbf{Z}_{Sup}^R(186)$. The datasets used in the weighted fine-tuning for \textbf{EE2} and \textbf{EE4} are those used for their component perception and control modules, i.e., no extra dataset is required for the our weighted end-to-end fine-tuning approach.

In the fine-tuning for \textbf{EE2} and \textbf{EE4}, the hyper-parameters such as $\beta$ and the percentage of real samples in a mini-batch for $L_p^{Sup}$ were empirically determined through 5-7 tuning experiments for each parameter. Too large or small $\beta$ or percentage of real samples could cause less improvement in hand-eye coordination or even worse performance.
The usage of real images for $L_p^{Sup}$ is crucial to avoid catastrophic forgetting of adapted perception modules. From Table~\ref{tab:e2e_experiments_settings}, we can see that \textbf{EE4} needs fewer real samples in a mini-batch than \textbf{EE2}. This is because $L_p^{Ad}$ in the fine-tuning for \textbf{EE4} provides extra help to avoid catastrophic forgetting.

\begin{figure*}[tb!]
\begin{center}
\includegraphics[width=1.9\columnwidth]{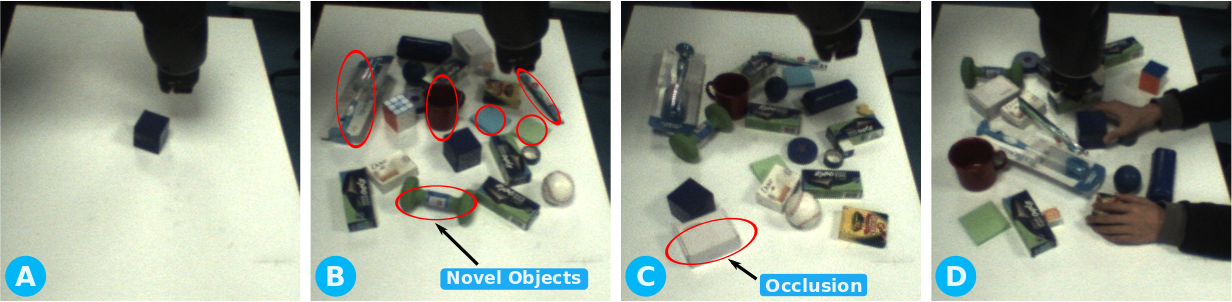}
\end{center}
\caption{Test cases for end-to-end performance. A: reaching the blue cuboid without distractor objects; B: reaching with seen and novel (not seen in training) objects as distractors; C: reaching with occlusion(s); D: reaching when the target is moving.}
\label{fig:test_cases}
\end{figure*}

\begin{table*}[tb!]
\caption{End-to-end Control Performance}
\label{tab:e2e_experiments_results}
\centering	
\renewcommand\arraystretch{1.2}
\begin{tabular}{c | c | c | c | c }	
\toprule 
Test Condition & Network / Method & $e_{med}$ [cm] & $e_{Q3}$ [cm] & $\sigma$ [\%] \\

\hline
\multirow{2}{*}{Control with ground-truth $\mathbf{\Theta}$} & The pseudo-inverse method  & {\bfseries 0.2} & {\bfseries 0.5} & {\bfseries 100} \\
\ & {\bfseries CC}  & 1.0 & 1.7 & {\bfseries 100}\\
\hline 
\multirow{5}{*}{Single object (Fig.~\ref{fig:test_cases}A)} &
{\bfseries Basline}: $\bold P_{\bold s1}$ + the pseudo-inverse method  & 2.5 & 3.7 & 86.7 \\
\ & {\bfseries EE0}: $\bold P_{\bold s1}$ + {\bfseries CC}  & 2.7 & 3.9 & 86.7\\
\ & {\bfseries EE1}: naively fine-tuned {\bfseries EE0} & 6.0 & 7.8 & 42.2\\
\ & {\bfseries EE2}: {\bfseries EE0} fine-tuned using our approach  & 
1.9 & 2.7 & 95.6\\
\ & {\bfseries EE3}: {\bfseries PP} + {\bfseries CC}  & 
2.1 & {\bfseries 2.6} & 95.6\\
\ & {\bfseries EE4}: {\bfseries EE3} fine-tuned using our approach  & 
{\bfseries 1.6} & 2.9 & {\bfseries 97.8}\\
\hline
\multirow{5}{*}{Clutter with novel objects (Fig.~\ref{fig:test_cases}B)} &
{\bfseries Baseline}  & 2.6 & 4.3 & 80.0 \\
\ & {\bfseries EE0}  & 3.5 & 4.8 & 68.9 \\
\ & {\bfseries EE1}  & 11.3 & 17.7 & 13.3 \\
\ & {\bfseries EE2}  & 2.1 & 2.7 & 95.6 \\
\ & {\bfseries EE3}  & 2.9 & 3.4 & 93.3 \\
\ & {\bfseries EE4}  & {\bfseries 1.8} & {\bfseries 2.6} & {\bfseries 97.8} \\
\hline
With occlusions (Fig.~\ref{fig:test_cases}C) & {\bfseries EE4}  & 4.6 & 8.5 & 48.9\\
\bottomrule 
\end{tabular}
\end{table*}

The baseline and combined networks were first evaluated in the real world without distractor objects on the table (Fig.~\ref{fig:test_cases}A), then the case with novel distractor objects in clutter (Fig.~\ref{fig:test_cases}B).
In the case of Fig.~\ref{fig:test_cases}B, 6 novel distractor objects (not seen in training) and 3 more white board eraser boxes (only the single box case was seen in training) were used in addition to those 11 distractor objects appeared in training. 
The metrics of control error ($e$) and success rate ($\sigma$) were used. $e_{med}$ and $e_{Q3}$ are the median and third quartile of control errors. Their results in 45 real-world reaching trials are listed in Table~\ref{tab:e2e_experiments_results}. The 45 trials were for the same targets and initial left arm configurations used in Section~\ref{sec:result_c}. The results for \textbf{CC} and the pseudo-inverse method (from Section~\ref{sec:result_c}) are also listed in the table (first two rows).

From Table~\ref{tab:e2e_experiments_results}, we can observe similar trends in the results for the cases of Fig.~\ref{fig:test_cases}A and Fig.~\ref{fig:test_cases}B. In comparison, the baseline and combined networks have larger errors in the case with distractor objects (more realistic).
In particular, \textbf{Baseline} achieved similar performances in the two test cases: similar median control errors (2.5\unit{cm} and 2.6\unit{cm}) and success rates (86.7\% and 80.0\%). Its errors are quite close to the perception error of $\bold P_{\bold s1}$ (2.8\unit{cm}, Q3:3.9\unit{cm}), but much larger than that of the pseudo-inverse method. This shows that the performance bottleneck of \textbf{Baseline} mainly comes from the perception module, and that $\bold P_{\bold s1}$ was generalized to both test cases.

In contrast, \textbf{EE0} achieved a similar performance in the case of Fig.~\ref{fig:test_cases}A (2.7\unit{cm} $e_{med}$ and 86.7\% $\sigma$), but has an obvious performance drop in the case of Fig.~\ref{fig:test_cases}B: median error and success rate decreased to 3.5\unit{cm} and 68.9\%. The decreased error is much larger than that of $\bold P_{\bold s1}$ and \textbf{CC}.
This shows that, for a directly connected network (\textbf{EE0}), the performance bottleneck comes from not only its component perception module ($\bold P_{\bold s1}$) but also the coordination between perception and control (\textbf{CC}).

A similar performance drop can also be observed from the results of \textbf{EE3} in the two test cases: $e_{med}$ decreased from 2.1\unit{cm} to 2.9\unit{cm}. However, the decrease of its success rate is trivial (from 95.6\% to 93.3\%). And \textbf{EE3} has much smaller errors and much higher success rates than \textbf{EE0} in both cases, although the two combined networks have the same control module \textbf{CC} and perception modules with similar errors ($\bold P_{\bold s1}$: 2.8\unit{cm}, Q3:3.9\unit{cm}; \textbf{PP}: 2.7\unit{cm}, Q3:3.9\unit{cm}).
These results show that a directly connected network (\textbf{EE3}) consisting of a perception module trained using ADT (\textbf{PP}) has better hand-eye coordination, i.e., \textbf{PP} has an output distribution that better fits the control module \textbf{CC} than $\bold P_{\bold s1}$.

After weighted end-to-end fine-tuning, both \textbf{EE2} and \textbf{EE4} achieved better performances than \textbf{EE0} and \textbf{EE3}. The improvement is significant, particularly in the case with novel distractor objects: \textbf{EE2} has 40.0\% smaller median control error and 38.8\% higher success rate; \textbf{EE4} has 37.9\% smaller median control error. The improvement of \textbf{EE4} in success rate is trivial, as \textbf{EE3} already has a very high success rate (93.3\%).
In contrast, \textbf{EE1} has a much worse performance than \textbf{EE0} in the two test cases (its performance in the case of Fig.~\ref{fig:test_cases}B is even worse).
This shows that our weighted end-to-end fine-tuning approach is able to significantly improve the performance of a combined network, but a naive approach could make the performance even worse. The end-to-end fine-tuning method works for both supervised perception adaptation (\textbf{EE2}) and adversarial discriminative transfer (\textbf{EE4}).

In addition, \textbf{EE2} and \textbf{EE4} even have much smaller control errors than the errors of their component perception modules ($\bold P_{\bold s1}$ and \textbf{PP}) in the challenging test case with novel distractor objects. If we individually evaluate the perception module in \textbf{EE2} using the same test set in Section~\ref{sec:result_supervised_p}, its perception error increased from 2.8 (Q3:3.9)\unit{cm} to 3.0 (Q3:5.2)\unit{cm}.
Similarly, the perception error of \textbf{EE4} increased from 2.7 (Q3:3.9)\unit{cm} to 4.2 (Q3:6.1)\unit{cm}.
These results indicate that the weighted end-to-end fine-tuning did improve the coordination between the perception and control modules (hand-eye coordination) in \textbf{EE2} and \textbf{EE4}, rather than improving them individually.

To further evaluate the performance of \textbf{EE4} in more challenging cases, we conducted more experiments with the target cuboid partially occluded, as shown in Fig.~\ref{fig:test_cases}C. From the results in the last row of Table~\ref{tab:e2e_experiments_results}, we can observe an obvious performance drop compared to that for the cases of Fig.~\ref{fig:test_cases}A and Fig.~\ref{fig:test_cases}B, but \textbf{EE4} was still able to reach half of the targets. We also tested \textbf{EE4} in the case when the target cuboid was moving (Fig.~\ref{fig:test_cases}D). It was able to adapt to target position changes in real time and performed well in most cases as shown in the attached video\footnote{The video is also on the project page \url{https://goo.gl/P1c354}.}.
\section{Discussion}
\label{sec:discussion}

The results described above lead us to the following observations:

\paragraph{Effectiveness of adversarial discriminative transfer}
The significant reduction (50\%) of required number of labelled real images for sim-to-real transfer of visuo-motor policies shows the effectiveness of the adversarial discriminative transfer. The PI controller and discriminator network architecture both play important roles in the approach. An acceptable transfer accuracy (3.0\unit{cm}) can be achieved with as few as 48 labelled real images, which is promising for robotic applications where labelling data is expensive or impractical. 

However, the approach in its current version can only effectively use a number of unlabelled real images no more than two times the number of labelled ones. This precludes using very few labelled and many unlabelled real images to further reduce the cost. More investigation is necessary to tackle this problem, enabling a few shots transfer of visuo-motor policies from simulation to the real world.

\paragraph{Value of a modular structure and end-to-end fine-tuning}

The significant performance improvement of \textbf{EE2} and \textbf{EE4} after end-to-end fine-tuning with weighted losses shows the effectiveness and scalability of the modular approach for more complicated tasks than the planar reaching~\citep{zhang2017acra}. Benefiting from the modular structure as well as the ADT approach, visuo-motor policies for a table-top reaching task can be learned and transferred from simulation to the real world with just 33225 simulated (including the 30225 ones for end-to-end fine-tuning) as well as 93 labelled and 186 unlabelled real samples, achieving a comparable performance to pure domain randomization approaches~\citep{james2017transferring} (the reaching stage of the multi-stage task) but with fewer training data in total. 

The modular approach can also be used in more general ways. Although we explicitly equated the bottleneck layer with the target object position in this work, the bottleneck in general could be any explicit or latent low-dimensional features (as in an auto-encoder). The perception and control modules can also be trained with other methods such as unsupervised learning and reinforcement learning. The effectiveness of the modular approach for reinforcement learning (DQN) has been validated in a planar reaching task~\citep{zhang2017acra,zhang2017cvpr}.

\paragraph{Domain randomization and adaptation}
In Section~\ref{sec:result_supervised_p}, the perception module trained with 3,000 simulated images (${\bold P}_{\bold s0}$) has a large error (47.1\unit{cm}), which is much higher than expected according to~\citep{tobin2017domain}. Apart from the experiments in Section~\ref{sec:result_supervised_p}, we also trained a number of perception modules using simulated images with random RGB values for the table, floor and robot body rather than $\pm10\%$ changes around the reference colors. However, this did not bring significant accuracy improvement. Possible reasons include: too simple textures (only random RGB values); too simple randomization for light conditions; no simulated shadows; or sensitivity to domain randomization parameters and tuning.
 
Nevertheless, with the ADT approach, the adaptation with just a few labelled real images (as few as 48) is able to transfer a network from simulation to the real world, and needs fewer simulated images than pure domain randomization approaches~\citep{tobin2017domain,james2017transferring}. The combination of domain randomization and adaptation is promising for more efficient deep neural network transfer.

\section{Conclusion}
\label{sec:conclusion}

In this paper, we proposed an adversarial discriminative transfer approach for cheaper transfer of visuo-motor policies from simulation to the real world. Its feasibility was demonstrated with a modular approach in the task of reaching a table-top object amongst clutter with a 7 DoF robotic arm in velocity mode. Our adversarial transfer approach reduced the labelled real data requirement by 50\%. Successful transfer was achieved with only 93 labelled and 186 unlabelled real images. By using weighted losses to fine-tune a combined network in an end-to-end fashion, its reaching accuracy was significantly improved (37.9\% better than that before fine-tuning), achieving a success rate of 97.8\% with a median control error of 1.8\unit{cm}. The learned policies are robust to novel distractor objects in clutter and even a moving target. The adversarial discriminative transfer along with the modular approach is promising for more efficient sim-to-real transfer of visuo-motor policies.
\section*{Acknowledgement}

This research was conducted by the Australian Research Council Centre of Excellence for Robotic Vision (project number CE140100016). 
This research has also been supported by the Australian Research Council's Linkage Infrastructure, Equipment and Facilities scheme, project number LE160100090. 
Additional computational resources and services were provided by the HPC and Research Support Group at QUT.


\bibliographystyle{SageH}
\bibliography{nc,fangyi}  

\end{document}